\documentclass{article}

\RequirePackage[OT1]{fontenc}
\RequirePackage{amsthm,amsmath}
\RequirePackage{natbib}
\RequirePackage[colorlinks,citecolor=blue,urlcolor=blue]{hyperref}

\usepackage[utf8]{inputenc}
\usepackage[T1]{fontenc}
\usepackage{natbib}
\usepackage{hyperref}
\usepackage{pgf}
\usepackage{import}
\usepackage{float}
\usepackage{amsmath,amssymb}
\usepackage{MnSymbol}
\usepackage{dsfont}
\usepackage{xcolor}
\usepackage{graphicx}
\usepackage{tikz}
\usepackage{indentfirst}
\usepackage{xspace}
\usepackage{algorithmic}
\usepackage{xfrac}
\usepackage{amsthm}
\usepackage[ruled,vlined]{algorithm2e}
\usepackage{subcaption}
\usepackage{framed}
\usepackage{bigints}
\usepackage{array}
\usepackage{caption}
\usepackage{tabularx}
\usepackage{multirow}
\usepackage{authblk}

\usetikzlibrary{backgrounds}
\usetikzlibrary{arrows.meta}

\newcommand\p[1]{{\left(#1\right)}}

\newcommand\brackets[1]{{\left[#1\right]}}

\newcommand\guill[1]{``#1''\xspace}

\usepackage{xstring}

\newcommand\llikli{\log{p}}
\newcommand\expectation{\mathbb{E}}

\newcommand\logistic{\operatorname{expit}}

\newcommand\norm[2]{\mathcal{N}\p{#1,#2}}
\newcommand\multinomial[2]{\mathcal{M}\p{#1;#2}}
\newcommand\bernoulli[1]{\mathcal{B}\p{#1}}
\newcommand\extbernoulli[2]{\mathcal{B}_{#1}\p{#2}}
\newcommand\categorial[2]{\operatorname{Cat}\p{#1,#2}}
\newcommand\NRX{q_{\gamma}}

\newcommand\iidsim{\overset{\textrm{iid}}\sim}
\newcommand\indsim{\overset{\textrm{ind}}\sim}
\newcommand\sachant[2]{\left.#1\middle|#2\right.}

\newcommand\Npi{\pi}
\newcommand\Npib{\boldsymbol{\pi}}

\newcommand\NXo{X^{(\textrm{o})}}
\newcommand\NXob{\boldsymbol{X}^{(\textrm{o})}}
\newcommand\NX{X}
\newcommand\NM{M}
\newcommand\NMb{\boldsymbol{M}}
\newcommand\NXb{\boldsymbol{X}}
\newcommand\NXc{X}
\newcommand\NXcb{\boldsymbol{X}}

\newcommand\NNA{\text{NA}}

\newcommand\Ni{i}
\newcommand\Nj{j}
\newcommand\Nq{q}
\newcommand\Nl{l}

\newcommand\Nnone{{n_1}}
\newcommand\Nntwo{{n_2}}
\newcommand\Nnq{{k_1}}
\newcommand\Nnl{{k_2}}

\newcommand\NA{A}
\newcommand\NB{B}
\newcommand\NP{C}
\newcommand\NQ{D}
\newcommand\NAb{\boldsymbol{A}}
\newcommand\NBb{\boldsymbol{B}}
\newcommand\NPb{\boldsymbol{C}}
\newcommand\NQb{\boldsymbol{D}}
\newcommand\NYone{Y}
\newcommand\NYonetild{\widehat{Y}}
\newcommand\NYonebtild{\widehat{\boldsymbol{Y}}}
\newcommand\NYoneb{\boldsymbol{Y}}
\newcommand\NYtwotild{\widehat{Z}}
\newcommand\NYtwobtild{\widehat{\boldsymbol{Z}}}
\newcommand\NYtwo{Z}
\newcommand\NYtwob{\boldsymbol{Z}}

\newcommand\Nmu{\mu}

\newcommand\Ntheta{\theta}
\newcommand\Nthetab{\boldsymbol{\theta}}
\newcommand\Nthetahat{\widehat{\Ntheta}}
\newcommand\NsigmaA{\sigma^2_{\NA}}
\newcommand\NsigmaB{\sigma^2_{\NB}}
\newcommand\NsigmaP{\sigma^2_{\NP}}
\newcommand\NsigmaQ{\sigma^2_{\NQ}}
\newcommand\Nalphaoneb{\boldsymbol{\alpha}}
\newcommand\Nalphatwob{\boldsymbol{\beta}}
\newcommand\Nalphaone{\alpha}
\newcommand\Nalphatwo{\beta}
\newcommand\Ngamma{\gamma}
\newcommand\NnuA{\nu^\p\NA}
\newcommand\NnuB{\nu^\p\NB}
\newcommand\NnuP{\nu^\p\NP}
\newcommand\NnuQ{\nu^\p\NQ}
\newcommand\NrhoA{\rho^\p\NA}
\newcommand\NrhoB{\rho^\p\NB}
\newcommand\NrhoP{\rho^\p\NP}
\newcommand\NrhoQ{\rho^\p\NQ}
\newcommand\Ntauone{\tau^\p\NYone}
\newcommand\Ntautwo{\tau^\p\NYtwo}

\newcommand\NnuAb{\boldsymbol\nu^\p\NA}
\newcommand\NnuBb{\boldsymbol\nu^\p\NB}
\newcommand\NnuPb{\boldsymbol\nu^\p\NP}
\newcommand\NnuQb{\boldsymbol\nu^\p\NQ}
\newcommand\NrhoAb{\boldsymbol\rho^\p\NA}
\newcommand\NrhoBb{\boldsymbol\rho^\p\NB}
\newcommand\NrhoPb{\boldsymbol\rho^\p\NP}
\newcommand\NrhoQb{\boldsymbol\rho^\p\NQ}
\newcommand\Ntauoneb{\boldsymbol\tau^\p\NYone}
\newcommand\Ntautwob{\boldsymbol\tau^\p\NYtwo}

\newcommand\Nodd{P_{\Ni\Nj}}
\newcommand\Noddp{P_{\Ni\Nj:\NXc_{\Ni\Nj}=1}}
\newcommand\Noddm{P_{\Ni\Nj:\NXc_{\Ni\Nj}=0}}

\newcommand\Kronecker[1]{\delta_{#1}}
\newcommand\Nlossitem[1]{l_{item}(#1)}
\newcommand\Nlossrow[1]{l_{row}(#1)}
\newcommand\Nlosscol[1]{l_{col}(#1)}
\newcommand\Prob[1]{p\p{#1}}

\newcommand\LBM{Latent Block Model\xspace}
\newcommand\LBMacro{LBM\xspace}
\newcommand\NMAR{Missing Not At Random\xspace}
\newcommand\NMARacro{MNAR\xspace}
\newcommand\MARacro{MAR\xspace}
\newcommand\MAR{Missing At Random\xspace}
\newcommand\MCARacro{MCAR\xspace}
\newcommand\MCAR{Missing Completely At Random\xspace}

\DeclareMathOperator*{\argmax}{arg\,max}
\DeclareMathOperator*{\argmin}{arg\,min}

\usepackage{xcolor}
\usepackage{ulem}
\usepackage{xcolor}
\usetikzlibrary{patterns}

\newtheorem{proposition}{Proposition}

\usepackage{enumitem}

\newif\ifgraph\graphfalse

\begin{document}

%\begin{frontmatter}

\title{Learning from missing data \\with the Latent Block Model}
\date{2020, October}

\author[1]{Gabriel Frisch}
\author[1]{Jean-Benoist Leger}
\author[1]{Yves Grandvalet}

\affil[1]{Université de technologie de Compiègne, CNRS, Heudiasyc (Heuristics
and Diagnosis of Complex Systems), CS 60 319 - 60 203 Compiègne Cedex}

\renewcommand\Affilfont{\itshape\small}

\maketitle

\begin{abstract}

Missing data can be informative. Ignoring this information can lead to misleading conclusions when the data model does not allow information to be extracted from the missing data.
We propose a co-clustering model, based on the \LBM, that aims to take advantage of this nonignorable nonresponses, also known as Missing Not At Random data (MNAR).
A variational expectation-maximization  algorithm is derived to perform inference and a model selection criterion is presented.
We assess the proposed approach on a simulation study, before using our model on the voting records from the lower house of the French Parliament, where our analysis brings out relevant groups of MPs and texts, together with a sensible interpretation of the behavior of non-voters.
\end{abstract}

\section{Introduction}
Biclustering or co-clustering simultaneously groups the rows and the columns of a data matrix. %into homogeneous blocks.
Co-clustering has found applications in many areas such as genomic analysis \citep{PONTES2015163, Kluger03spectralbiclustering}, text analysis \citep{Dhillon, SELOSSE2020107315}, collaborative filtering \citep{recsyscoclustering, shanbanerjee}, or political analysis \citep{Latouche_2011,wyse2012block}.
Co-clustering methods can be divided into categories such as, but not limited to, spectral methods \citep{dhillon2001cocluster, Kluger03spectralbiclustering}, mutual information methods \citep{Dhillon}, modularity based methods \citep{labiod2011}, non negative matrix tri-factorization \citep{trinmf} or model-based methods.
Among the model-based methods, the Latent Block Model \citep{lbmgg, govaert2010, lomet2012, keribin:hal-00802764} relies on mixtures, assuming that the observations are generated from finite mixture components in rows and columns.
    
Most standard methods of clustering or co-clustering presuppose complete information and cannot be applied with missing data, or may provide misleading conclusions when missingness is informative.
A careful examination of the data generating process is necessary for the processing of missing values, which requires identifying the type of missingness \citep{rubin}: Missing Completely At Random (\MCARacro) refers to the mechanism in which the probability of being missing does not depend on the variable of interest or any other observed variable; whereas in Missing At Random (MAR) the probability of being missing depends on some observed data but is still independent from the non-observed data; and finally Missing Not At Random (\NMARacro) refers to the mechanism in which the probability of being missing depends on the actual value of the missing data.
Under the MAR hypothesis, no information on the generation of data can be extracted from its absence, but under a \NMARacro assumption, this absence is informative, and ignoring this information in likelihood-based imputation methods may lead to strong biases in estimation \citep{little2019statistical}. 
Missing Not At Random is also known as non-ignorable missingness, in opposition to the ignorable missingness of \MCARacro and MAR settings, as the absence of data is assumed to convey some information.

In this paper, we aim at clustering the rows and columns of a data matrix whose entries are missing not at random. Equivalently, we consider the clustering of the vertices of a bipartite graph whose edges are missing not at random. For this purpose, we introduce  a  co-clustering  model that combines a \NMARacro missingness model with the Latent Block Model (LBM).

In Section \ref{sec:lbm} we review the Latent Block Model introduced by \cite{lbmgg}. 
In Section \ref{sec:missingnessmodel}, we introduce our model, a LBM extended to a \NMARacro missingness process, and propose, in Section \ref{sec:inference}, a variational EM algorithm to infer its parameters. 
We also introduce, in Section \ref{sec:modelselection}, an Integrated Completed Likelihood (ICL) criterion to tackle model selection.
We then conduct  experiments  on  synthetic datasets in Section \ref{sec:simulatedata} to show that the overall approach is relevant to co-cluster \NMARacro data.
Finally, an analysis of the voting records of the lower house of the French Parliament is presented in Section \ref{sec:realdata}.

\subsection{Related Works}
Up to our knowledge, all existing co-clustering methods consider that missing data is either \MCARacro or MAR \citep{SELOSSE2020106866, jacques:hal-01448299, Papalexakis}, except one proposed by \cite{corneli:hal-01978174} used to co-cluster ordinal data. 
Their model is very parsimonious as it assumes that both data and missingness are only dependent on the row and column clusters. 
In this setting, they are able to consider \NMARacro data even if they suppose that missingness depends indirectly from the value of the data.
The model we propose is less parsimonious, thus more flexible, as it supposes that missingness depends both on the value of the data and on the row and column indexes (not only on their respective cluster indexes).
In addition to that, our missing data model can be easily re-used for any other statistical co-clustering model as it is weakly-coupled to the generative model of the full data matrix.

In the simple clustering framework, few mixture models handling \NMARacro data have been proposed.
\citet{Marlin} combine a multinomial mixture clustering model, used as a complete data model, with a missingness model of type \NMARacro.
They propose two versions of their missingness model.
The first one, called CPT-v, models the data observation probability depending only on the underlying value of the data.
The second one, called Logit-vd, allows the probability of a data entry to be missing to depend both on the value of the underlying data and the characteristics of the column, giving more flexibility to the model.
Our missingness model respects the symmetry of the co-clustering problem by depending identically on the characteristics of the row and column.
\citet{Kim} propose Bayesian-BM/OR, a simple mixture model of binomials in a Bayesian formalism.
The \NMARacro missingness is modeled by three factors, related to the row, the column and the data value, all three being modeled by Bernoulli variables combined together by a ``or'' logical operator.
The choice of this missingness model is motivated by algorithmic considerations that are not relevant for co-clustering models.
\citet{tabouy}, in a graph perspective, deal with nonobserved dyads during the sampling of a network and consecutive issues in the inference of the stochastic block model.
They propose three different \NMARacro sampling designs in which observing dyads depends either on their underlying value, or on the class or on the degree of the nodes.
The Stochastic Block Model, though similar from the \LBM we use, is not usable for co-clustering purposes.

Also related to missing data but not to clustering, \NMARacro is also considered in the Matrix Factorization framework.
\citet{Steck} derives a weighted MF model and optimizes the parameters based on a metric that is robust to \NMARacro data.
\citet{hernandez} use a double probabilistic MF model; one is for the complete data and one for the missing data, where users and items propensities are both modeled with low rank matrices.
\citet{schnabel2016recommendations} propose an empirical risk minimization framework to derive a propensity scored matrix factorization method that can account for selection bias.
\section{The \LBM}
\label{sec:lbm}
    The \LBM (\LBMacro) is a {\em co-clustering} model that classifies jointly the rows and the columns of a data matrix \citep{lbmgg}.
    This probabilistic generative model assumes a double partition on the rows and the columns  of a $(\Nnone \times \Nntwo)$ data matrix  $\NXb$ that corresponds to a strong structure of the matrix in homogeneous blocks.
    This structure is unveiled by reordering the rows and columns according to their respective cluster index; for $\Nnq$ row clusters and $\Nnl$ column clusters, the reordering reveals $\Nnq\times\Nnl$ homogeneous blocks in the data matrix.
    Note that we adopt here the original view where the data matrix is interpreted as a data table. 
    The binary matrix $\NXb$ can also be interpreted as the biadjacency matrix of a bipartite graph, whose two sets of vertices corresponds to the rows and columns of the data matrix.
    In this interpretation, $\NX_{\Ni\Nj} = 1$ if an edge is present between ``row node'' $\Ni$ and ``column node'' $\Nj$, and $\NX_{\Ni\Nj} = 0$ otherwise.

    \ifgraph 
    As firstly introduced, the \LBMacro considers that $\NXb$ is random binary matrix.
    From a graph inference point of view, this binary matrix $\NXb$ ($\Nnone \times \Nntwo$) is the adjacency matrix of a bipartite graph with $\Nnone$ nodes of type (1) and $\Nntwo$ nodes of type (2).
    The random variable $\NX_{\Ni\Nj}$ is associated for each pair of nodes ($\Ni$,$\Nj$) respectively of type (1) and type (2) coding the presence or the absence of edge between $\Ni$ and $\Nj$.
    $\NX_{\Ni\Nj} = 1$ if the edge is present and $\NX_{\Ni\Nj} = 0$ otherwise.
    \fi
     For the $(\Nnone \times \Nntwo)$ data matrix $\NXb$, two partitions are defined by the latent variables $\NYoneb$ and $\NYtwob$, with $\NYoneb$ being the $\Nnone \times \Nnq$ indicator matrix of the latent row clusters
    ($\NYone_{\Ni\Nq} = 1$ if row $\Ni$ belongs to group $\Nq$ and $\NYone_{\Ni\Nq} = 0$ otherwise),
     and $\NYtwob$ being the $\Nntwo \times \Nnl$ indicator matrix of the latent column cluster.
     The group indicator of row $\Ni$ will be denoted  $\NYoneb_\Ni$, and 
    similarly, the group indicator of column $\Nj$ will be denoted $\NYtwob_\Nj$.
    The \LBMacro makes several assumptions on the dependencies:
	\paragraph{Independent rows and column clusters} The latent variables $\NYoneb$ and $\NYtwob$ are \textit{a priori} independent.
	\begin{equation*}
	p(\NYoneb, \NYtwob) = p(\NYoneb)p(\NYtwob)
	\enspace.
	\end{equation*}
	Note that \textit{a priori} independence does not imply \textit{a posteriori} independence: given the data matrix $\NXb$, the two partitions are (hopefully) not independent.
	% \[p(\NYoneb, \NYtwob|\NXb) \neq p(\NYoneb|\NXb)p(\NYtwob|\NXb) \enspace.\]
	%
	\paragraph{Independent and identically distributed row clusters}
	 The latent variables $\NYoneb$ are independent and follow a multinomial distribution $\multinomial1{\Nalphaoneb}$, where $\Nalphaoneb=(\Nalphaone_1,...,\Nalphaone_\Nnq )$ is the mixing proportions of rows:
	\begin{align*}
		&p(\NYoneb; \Nalphaoneb) = \prod_\Ni{p(\NYoneb_\Ni; \Nalphaoneb)} \\
		&p(\NYone_{\Ni\Nq}=1; \Nalphaoneb) = \Nalphaone_{\Nq}
		\enspace,
	\end{align*}
	with $\Nalphaoneb \in S_{(\Nnq-1)}=\{\Nalphaoneb\in\mathbb{R}_+^{\Nnq}| \sum_\Nq{\Nalphaone_\Nq=1}\}$.
	\paragraph{Independent and identically distributed column clusters}
	Likewise, the latent variables $\NYtwob$ are independent and follow a multinomial distribution $\multinomial1{\Nalphatwob}$, where $\Nalphatwob~=~(\Nalphatwo_1,...,\Nalphatwo_\Nnl )$ is the mixing proportions of columns:
	\begin{align*}
		&p(\NYtwob; \Nalphatwob) = \prod_\Nj{p(\NYtwob_\Nj; \Nalphatwob)} \\
		&p(\NYtwo_{\Nj\Nl}=1; \Nalphatwob) = \Nalphatwo_{\Nl}
		\enspace,
	\end{align*} with $\Nalphatwob \in S_{(\Nnl-1)}$.
	\paragraph{Given row and column clusters, independent and identically distributed block entries} Given the row and colum clusters $(\NYoneb,\NYtwob)$, the entries $\NX_{\Ni\Nj}$ are independent and follow a Bernoulli distribution of parameter $\Npib=(\Npi_{\Nq\Nl};\Nq=1,...,\Nnq; \Nl=1,...,\Nnl)$: all elements of a block follow the same probability distribution.
	\begin{align*}
		&p(\sachant{\NXb}{\NYoneb,\NYtwob; \Npib}) = \prod_{\Ni\Nj}{p\p{\sachant{\NX_{\Ni\Nj}}{\NYoneb_\Ni,\NYtwob_\Nj; \Npib}}} \\
		&p\p{\sachant{\NX_{\Ni\Nj}=1}{\NYone_{\Ni\Nq}\NYtwo_{\Nj\Nl}=1; \Npib}} = \Npi_{\Nq\Nl} 
		\enspace.
	\end{align*}

    To summarize, 
    the parameters of the \LBMacro are $\theta = (\Nalphaoneb, \Nalphatwob, \Npib)$ and the probability mass function of $\NXb$ can be written as:
    \begin{equation*}
    p\p{\NXb; \theta} = \sum_{(\NYoneb,\NYtwob) \in I \times J}{ \p{\prod_{\Ni,\Nq}{{\Nalphaone_\Nq}^{\NYone_{\Ni\Nq}}}} \p{\prod_{\Nj,\Nl}{{\Nalphatwo_\Nl}^{\NYtwo_{\Nj\Nl}}}} \p{\prod_{\Ni,\Nj,\Nq,\Nl}{\phi(\NX_{\Ni\Nj}}; \Npi_{\Nq\Nl})^{\NYone_{\Ni\Nq} \NYtwo_{\Nj\Nl} }} }
    \enspace,
    \end{equation*}
    where {$\phi(\NX_{\Ni\Nj}; \Npi_{\Nq\Nl}) = \Npi^{\NX_{\Ni\Nj}}_{\Nq\Nl}(1-\Npi_{\Nq\Nl})^{1-\NX_{\Ni\Nj}}$} is the mass function of a Bernoulli variable and where $I$ (resp. $J$) denotes the set of all possible partitions of rows (resp. columns) into $\Nnq$ (resp. $\Nnl$) groups.

    \begin{figure}[tb]
        \begin{framed}
        \centering
        \begin{minipage}{.3\textwidth}
              \centering
              \begin{tikzpicture}[scale=1, every node/.style={scale=1.3}]
                 \node[circle, draw=black] (Y1) at (0,0) {$\NYoneb_\Ni$};
                 \node[circle,draw=black] (Y2) at (2,0) {$\NYtwob_\Nj$};
                 \node[circle,draw=black, inner sep=1pt] (X) at (1,-1.5) {$\NX_{\Ni\Nj}$};
                 \draw[-{Latex[length=3mm, width=2mm]}] (Y1) -- (X);
                 \draw[-{Latex[length=3mm, width=2mm]}] (Y2) -- (X);
              \end{tikzpicture}
        \end{minipage}
        \begin{minipage}{.6\textwidth}
              \[
                \left\{
                  \begin{array}{l}
                    \forall \Ni,\;\NYoneb_\Ni \iidsim \multinomial1{\Nalphaoneb} \\
                    \forall \Nj,\; \NYtwob_\Nj \iidsim \multinomial1{\Nalphatwob}\\
                    \forall \Ni,\Nj, \;\sachant{\NX_{\Ni\Nj}}{\NYone_{\Ni\Nq}=1, \NYtwo_{\Nj\Nl} = 1} \;\indsim\;\extbernoulli{}{\Npi_{\Nq\Nl}}
                  \end{array}
                \right.
              \]
              \[
              \text{with } \Nalphaoneb \in \mathbf{S}_{\Nnq-1},\; \Nalphatwob \in \mathbf{S}_{\Nnl-1} \text{ and } \Npi_{ql} \in [0,1]
              \]
        \end{minipage}
        \label{fig:graphiclbmbernouilli}

    \end{framed}
    \captionof{figure}{Summary of the standard \LBM with binary data.}
    \end{figure}
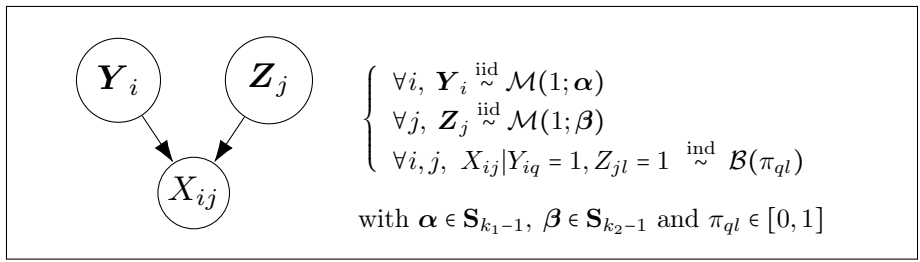

    \section{Extension to Informative Missing Data}
    \label{sec:missingnessmodel}
    The standard \LBM does not accommodate missing observations, that is, the data matrix $\NXb$ is fully observed.
    %. In other words, in the binary case, any $\NX_{\Ni\Nj}$ is either 1 or 0 and 
    %We propose, an extension of the \LBMacro that will be able to take into account and to estimate the missing data. 
    This section introduces our missingness model, which will be coupled to the \LBMacro, thereby enabling to process missing data.

    We start by introducing some notation: from now on, $\NXob$ will denote the \guill{partially observed} data matrix, with missing entries, whereas $\NXcb$ denotes the \guill{full} (unobserved) data matrix, without missing entries. The partially observed matrix $\NXob$ is identical to the full matrix $\NXcb$ except for the missing entries;
    $\NXob$ takes its values in $\{0,1, \NNA\}$, where $\NNA$ denotes a missing value.
    It will be convenient to introduce a binary mask matrix $\NMb$ that indicates the {\em non-missing} entries of $\NXob$: if $\NM_{\Ni\Nj}=0$, then $\NXo_{\Ni\Nj}=\NNA$.

    \subsection{Models of Missingness}
    
    % First explain what the models of missingness are
    The three main types of missingness are \MCAR (\MCARacro), \MAR (\MARacro), and \NMAR (\NMARacro). 
    We propose here a model for each missingness type.
    Instead of directly modeling the probability of being missing, we will model a real variable that defines the log-odds of this probability. This log-odds will be called here the ``propensity'' to be missing.
    
    \iffalse
    When facing \NMARacro data, most existing missingness models are making an \MCARacro or \MARacro assumption for computational simplicity reasons during the inference process. 
    Indeed any generative model combined with a \MCARacro or \MARacro missingness model can be trained separately as the likelihood of the overall model is factorizable between the two models.
    Under the \MCARacro and \MARacro hypothesis, no information on the generation of data can be extracted from an absence of data, but under a \NMARacro assumption, this absence is informative, and ignoring this information may lead to strong biases in estimations that may in turn drastically affect the  classifications. 
    From these considerations, we propose an extension of the \LBMacro model enabling to deal with Not Missing At Random missingness data. 
    The resulting model is made up of two distinct parts: the \LBMacro used to model the \guill{full} data matrix and the \NMARacro model.
    The two models are merged together and can not be trained separately because of their mutual dependency; however for clarity reasons, we first describe the model of the missingness process and then its use with the \LBMacro.
    \fi 
          
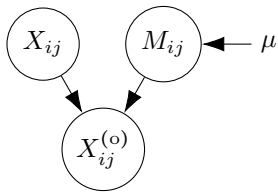
\begin{figure}
        \centering
        \begin{tikzpicture}[scale=0.8, every node/.style={scale=1}]
                    \node (mu) at (1.75,2) {$\Nmu$};
                    \node[circle,draw=black] (M) at (0,2) {${M}_{\Ni\Nj}$};
                    \node[circle,draw=black] (Xc) at (-2,2) {$\NXc_{\Ni\Nj}$};
                    \node[circle,draw=black, inner sep=0.5ex] (X) at (-1,0.25) {$\NXo_{\Ni\Nj}$};
                    \draw[{Latex[length=3mm, width=2mm]}-] (M) -- (mu);
                    \draw[{Latex[length=3mm, width=2mm]}-] (X) -- (Xc);
                    \draw[{Latex[length=3mm, width=2mm]}-] (X) -- (M);
                    %\node[rectangle,draw=white, scale=0.7] at (-0.3,-2.6) {\LBMacro};
                    \end{tikzpicture}
                \caption{\label{CMARmodel}Graphical representation of the \MCARacro model. 
                The partially observed entry $\NXo_{\Ni\Nj}$ is generated by the corresponding entries of (i) the full matrix $\NXc_{\Ni\Nj}$ and (ii) the binary mask $\NM_{\Ni\Nj}$.
                The binary mask $\NMb$ does not depend on $\NXcb$ and is defined here from a single global effect parameter $\Nmu$.}
    \end{figure}

    %
    \paragraph{\MCAR (\MCARacro)} Missingness does not depend on data, whether observed or not.
    A simple model of missingness is obtained by assuming that every entry of $\NXob$ has the same propensity of being missing. 
    This is modeled by a single propensity parameter $\Nmu$.
    The graphical representation of this model is shown in Figure~\ref{CMARmodel}.

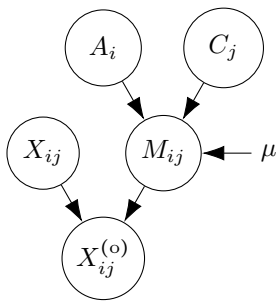
\begin{figure}
                \centering
                \begin{tikzpicture}[scale=0.8, every node/.style={scale=1}]
                    \node (mu) at (1.75,2) {$\Nmu$};
                    \node[circle,draw=black] (M) at (0,2) {${M}_{\Ni\Nj}$};
                    \node[circle,draw=black] (Xc) at (-2,2) {$\NXc_{\Ni\Nj}$};
                    \node[circle,draw=black, inner sep=0.5ex] (X) at (-1,0.25) {$\NXo_{\Ni\Nj}$};
                    \node[circle,draw=black, inner sep=1.25ex] (A) at (-1,3.75) {$\NA_\Ni$};
                    \node[circle,draw=black, inner sep=1.25ex] (P) at (1,3.75) {$\NP_\Nj$};
                    \draw[{Latex[length=3mm, width=2mm]}-] (X) -- (Xc);
                    \draw[{Latex[length=3mm, width=2mm]}-] (M) -- (A);
                    \draw[{Latex[length=3mm, width=2mm]}-] (M) -- (P);
                    \draw[{Latex[length=3mm, width=2mm]}-] (M) -- (mu);
                    \draw[{Latex[length=3mm, width=2mm]}-] (X) -- (M);
                \end{tikzpicture}
                \caption{\label{MARmodel}
                    Graphical representation of the \MARacro model. 
                    The partially observed entry $\NXo_{\Ni\Nj}$ is generated by the corresponding entries of (i) the full matrix $\NXc_{\Ni\Nj}$ and (ii) the binary mask $\NM_{\Ni\Nj}$. 
                    The binary mask $\NMb$ does not depend on $\NXcb$ and is and is defined by a global effect parameter $\Nmu$ and two latent variables $\NAb$ and $\NPb$ that enable deviations from $\Nmu$.}
            \end{figure}

    \paragraph{\MAR (\MARacro)} Missingness depends on the observed data, but not on the unobserved data.
            The previous missingness model can be enlarged by allowing the propensity of missingness to depend on the row and column indexes.
            To do so, we can introduce a latent variable for every row, denoted $\NAb$, and another one for every column, denoted $\NPb$.
            For the sake of simplicity, all latent variables $\NA_\Ni$ and $\NP_\Nj$ are assumed independent.
            They allow deviations from the global propensity $\Nmu$.
            The graphical representation of this model is shown in Figure~\ref{MARmodel}.

\begin{figure}
				\centering
                \begin{tikzpicture}[scale=0.8, every node/.style={scale=1}]
                    \node[circle,draw=black, inner sep=1.25ex] (A) at (-2.2,3.75) {$\NA_\Ni$};
                    \node[circle,draw=black, inner sep=1.25ex] (B) at (-0.8,3.75) {$\NB_\Ni$};
                    \node[circle,draw=black, inner sep=1.25ex] (P) at (0.8,3.75) {$\NP_\Nj$};
                    \node[circle,draw=black, inner sep=1.25ex] (Q) at (2.2,3.75) {$\NQ_\Nj$};
                    \node (mu) at (1.75,2) {$\Nmu$};
                    \node[circle,draw=black] (Xc) at (-2,2) {$\NXc_{\Ni\Nj}$};
                    \node[circle,draw=black, inner sep=0.5ex] (X) at (-1,0.25) {$\NXo_{\Ni\Nj}$};
                    \draw[{Latex[length=3mm, width=2mm]}-] (X) -- (Xc);
                    % \node[circle,draw=black, inner sep=2.25ex] (Xc) at (0,-2) {};
                    \node[circle,draw=black] (M) at (0,2) {$M_{\Ni\Nj}$};
                    \draw[{Latex[length=3mm, width=2mm]}-] (M) -- (Xc);
                    \draw[{Latex[length=3mm, width=2mm]}-] (M) -- (A);
                    \draw[{Latex[length=3mm, width=2mm]}-] (M) -- (B);
                    \draw[{Latex[length=3mm, width=2mm]}-] (M) -- (P);
                    \draw[{Latex[length=3mm, width=2mm]}-] (M) -- (Q);
                    \draw[{Latex[length=3mm, width=2mm]}-] (M) -- (mu);
                    \draw[{Latex[length=3mm, width=2mm]}-] (X) -- (M);
                \end{tikzpicture}
                \caption{\label{MNARmodel}
                    Graphical representation of the \NMARacro model. 
                    The partially observed entry $\NXo_{\Ni\Nj}$ is generated by the corresponding entries of (i) the full matrix $\NXc_{\Ni\Nj}$ and (ii) the binary mask $\NM_{\Ni\Nj}$. 
                    The binary mask $\NMb$ depends on $\NXcb$ and is defined by a global effect parameter $\Nmu$, two latent variables $\NAb$ and $\NPb$ that enable deviations from $\Nmu$, and two latent variables $\NBb$ and $\NQb$, which drive the deviations from the MAR model.}
            \end{figure}
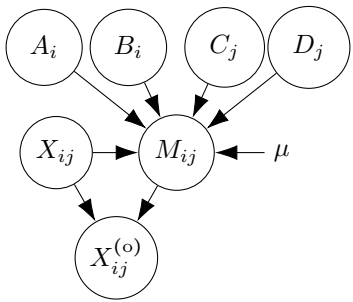

            \paragraph{\NMAR (\NMARacro)}
            Missingness here depends on unobserved data:
            the probability of observing the entries of the matrix depends on their values, whether observed or not.
            We equip the previous model with two additional latent variables to adapt the propensity of each entry of the data matrix to the unobserved data, that is, to $\NXc_{\Ni\Nj}$.
            These new row and column latent variables, $\NBb$ and $\NQb$, adjust the propensity of missingness according to the actual value of $\NXc_{\Ni\Nj}$.
            The graphical representation of this model is shown in Figure~\ref{MNARmodel}.

        %We propose a missingness model of type \NMARacro based on the logistic model.
        %In the logistic model, the log-odds is a linear combination of variables.
        We model the latent variables $\NAb$, $\NBb$, $\NPb$, and $\NQb$ with Gaussian distributions centered at zero with free variances $\NsigmaA$, $\NsigmaB$, $\NsigmaP$, and $\NsigmaQ$, respectively:
        \begin{equation*}
          \left\{
            \begin{array}{l}
                \forall \Ni,\qquad \NA_\Ni \iidsim \norm0{\NsigmaA} ,\quad \NB_\Ni \iidsim \norm0{\NsigmaB}\\
                \forall \Nj,\qquad \NP_\Nj \iidsim \norm0{\NsigmaP} ,\quad \NQ_\Nj \iidsim \norm0{\NsigmaQ}
            \end{array}
          \right.
          \enspace.
        \end{equation*}
        The global parameter $\Nmu$ and the latent variables define the propensity of missingness, that is, the log-odds of being missing as follows:
        \begin{equation}
            \forall \Ni, \Nj \quad
            \Nodd =\left\{
            \begin{array}{l}
                \Nmu+\NA_\Ni +\NB_\Ni+\NP_\Nj+\NQ_\Nj \quad \text{if} \quad \NXc_{\Ni\Nj}=1  \\
                \Nmu+\NA_\Ni -\NB_\Ni+\NP_\Nj-\NQ_\Nj \quad \text{if} \quad \NXc_{\Ni\Nj}=0  \\
            \end{array}
            \right.
            \enspace.
        \end{equation}
        Then, given this  propensity, every element $\NM_{\Ni\Nj}$ of the mask matrix is independent and follows a Bernoulli distribution:
        \begin{equation}
            \forall \Ni, \Nj \quad
                \sachant{ M_{\Ni\Nj} }{\NA_\Ni,\NB_\Ni,\NP_\Nj,\NQ_\Nj, \NXc_{\Ni\Nj} } \; \indsim \;  \bernoulli{\logistic\p{\Nodd}}
            \enspace,
        \end{equation}
        with {$\logistic(x)= 1/(1+\exp(-x))$}.

        Note that, if we omit the latent variables $\NB_\Ni$ and $\NQ_\Nj$, the missingness model follows the \MARacro assumption since $\Nodd$, and thus $M_{\Ni\Nj}$, is then independent of $\NXc_{\Ni\Nj}$.
        If we also omit the latent variables $\NA_\Ni$ and $\NP_\Nj$, the missingness model follows the \MCARacro assumption.

        This model of missingness can be used for several applications.
        One of these, collaborative filtering, uses the history of user ratings to build a recommendation system.
        For this application, an \MCARacro modeling  means that the probability of observing a rating for a particular item does not depend on the user nor the item;
        an \MARacro modeling  means  that missingness can depend on the user or the item;  for example, some people give their opinion more often than others.
        The \MARacro simplifying assumption is often used in collaborative filtering.
        However, \cite{Marlin07} show that there is often a dependency between the rating frequency and the underlying preference level, lending support to the hypothesis that ratings are generated by a MNAR process, where missingness depends on the actual rating that would be given.
        %A MNAR modelization of the missing data would mean that the probability of observing a rating for a particular item depends on the user's rating for that item.
        Some people give their opinion more often when they are satisfied and other ones when they are dissatisfied.
        Most collaborative filtering methods do not have a principled method for extracting information from missing data, which can lead to strong biases in estimations that may in turn drastically affect predictions  \citep{hernandez}.
        Our missingness model allows to account for  the users' propensity to give their opinion, and for the items' propensity to be rated, that is, their notoriety.
        These propensities could also reflect exogenous factors such as price; for example, more expensive items could be evaluated more often.

        \subsection{\LBMacro with MNAR data}
        \label{sect:lbmextended}
        We extend the standard $\LBMacro$ using the previous modeling to \NMARacro data. 
        
\begin{figure}
            \begin{framed}
            \centering
            \begin{minipage}{1.\textwidth}
                  \centering
                        \begin{tikzpicture}[scale=0.8, every node/.style={scale=1}]
                        	\node[circle, draw=black] (Y1) at (0,0) {$\NYoneb_\Ni$};
                        	\node[circle,draw=black] (Y2) at (2,0) {$\NYtwob_\Nj$};
                        	\node[circle,draw=black] (A) at (4,0) {$\NA_\Ni$};
                        	\node[circle,draw=black] (B) at (6,0) {$\NB_\Ni$};
                          \node[circle,draw=black] (P) at (8,0) {$\NP_\Nj$};
                        	\node[circle,draw=black] (Q) at (10,0) {$\NQ_\Nj$};
                            \draw[color=black!30!white] (-1,1) rectangle (3,-3);
                            \node[scale=0.7, color=black!30!white] at (-0.58,-2.8) {\LBMacro};
                            \draw[color=black!30!white, dashed] (0,-3) --  (0,-1) -- (3.3,-1) -- (3.3,1) -- (10.8,1)  -- (10.8,-3) -- (0,-3);
                            \node[scale=0.7, color=black!30!white] at (9.,-2.81) {MNAR missingness model};
                           \node[circle,draw=black] (Xc) at (1,-2) {$\NXc_{\Ni\Nj}$};
                        	\node[circle,draw=black] (M) at (7,-2) {$M_{\Ni\Nj}$};
                        	\node[circle,draw=black] (Xo) at (4,-4.5) {$\NXo_{\Ni\Nj}$};
                        	\draw[-{Latex[length=3mm, width=2mm]}] (Y1) -- (Xc);
                        	\draw[-{Latex[length=3mm, width=2mm]}] (Y2) -- (Xc);
                        	\draw[-{Latex[length=3mm, width=2mm]}] (A) -- (M);
                        	\draw[-{Latex[length=3mm, width=2mm]}] (B) -- (M);
                          \draw[-{Latex[length=3mm, width=2mm]}] (P) -- (M);
                        	\draw[-{Latex[length=3mm, width=2mm]}] (Q) -- (M);
                           \draw[-{Latex[length=3mm, width=2mm]}] (Xc) -- (Xo);
                           \draw[-{Latex[length=3mm, width=2mm]}] (Xc) -- (M);
                        	\draw[-{Latex[length=3mm, width=2mm]}] (M) -- (Xo);
                        \end{tikzpicture}
            \end{minipage}\\\bigskip
            \begin{minipage}{1.\textwidth}
                  % \centering
                  \begin{align}
                    % \forall \Ni,\Nj, \left\{
                        &\text{Latent Block Model} \nonumber\\
                        &\quad \NYoneb_\Ni \iidsim \multinomial1{\Nalphaoneb}, \qquad  \Nalphaoneb \in \mathbf{S}_{\Nnq-1} \nonumber\\
                        &\quad \NYtwob_\Nj \iidsim \multinomial1{\Nalphatwob}, \qquad  \Nalphatwob \in \mathbf{S}_{\Nnl-1} \nonumber\\
                        &\quad \p{\sachant{\NXc_{\Ni\Nj}}{\NYone_{\Ni\Nq}=1, \NYtwo_{\Nj\Nl} = 1}} \indsim \extbernoulli{}{\Npi_{\Nq\Nl}}, \qquad  \Npi_{ql} \in (0,1) \nonumber\\
                        &\text{MNAR model} \nonumber\\
                        &\quad\NA_\Ni \iidsim \norm0{\NsigmaA}, \qquad  \NsigmaA \in \mathds{R}_{+}^{*} \nonumber\\
                        &\quad\NB_\Ni \iidsim \norm0{\NsigmaB}, \qquad  \NsigmaB \in \mathds{R}_{+}^{*} \nonumber\\
                        &\quad\NP_\Nj \iidsim \norm0{\NsigmaP}, \qquad  \NsigmaP \in \mathds{R}_{+}^{*} \nonumber\\
                        &\quad\NQ_\Nj \iidsim \norm0{\NsigmaQ}, \qquad  \NsigmaQ \in \mathds{R}_{+}^{*} \nonumber\\
                        &\quad\p{\sachant{ M_{\Ni\Nj} }{\NA_\Ni,\NB_\Ni,\NP_\Nj,\NQ_\Nj, \NXc_{\Ni\Nj}=1 }} \indsim  \bernoulli{\logistic\p{\Nmu+\NA_\Ni +\NB_\Ni+\NP_\Nj+\NQ_\Nj}} \nonumber\\
                        &\quad\p{\sachant{ M_{\Ni\Nj} }{\NA_\Ni,\NB_\Ni,\NP_\Nj,\NQ_\Nj, \NXc_{\Ni\Nj}=0 }} \indsim  \bernoulli{\logistic\p{\Nmu+\NA_\Ni -\NB_\Ni+\NP_\Nj-\NQ_\Nj}} \nonumber\\
                        % \sachant{ M_{\Ni\Nj} }{\NA_\Ni,\NB_\Ni,\NP_\Nj,\NQ_\Nj, \NXc_{\Ni\Nj}}
                        % &\indsim
                        % \left\{
                        %   \begin{array}{cll}
                        %     \bernoulli{\logistic\p{\}}   &\text{ if } \quad \NXc_{\Ni\Nj}=1 \\
                        %      \bernoulli{\logistic\p{\Noddm}}  &\text{ if } \quad \NXc_{\Ni\Nj}=0 \\
                        %   \end{array}
                        % \right. \nonumber\\
                        &\text{Observations are generated according to:} \nonumber\\
                        % &\p{\sachant{ \NXo_{\Ni\Nj} }{ \NXc_{\Ni\Nj}, M_{\Ni\Nj}=1}} = \NXc_{\Ni\Nj} \nonumber\\
                        % &\p{\sachant{ \NXo_{\Ni\Nj} }{ \NXc_{\Ni\Nj}, M_{\Ni\Nj}=0}} = \NNA \nonumber
                        &\quad\p{\sachant{ \NXo_{\Ni\Nj} }{ \NXc_{\Ni\Nj}, M_{\Ni\Nj}}} =
                        \left\{
                          \begin{array}{cll}
                            \NXc_{\Ni\Nj} &\text{ if } \quad \NM_{\Ni\Nj} = 1 \\
                            \NNA  &\text{ if } \quad \NM_{\Ni\Nj} = 0 \\
                          \end{array}
                        \right. \nonumber
                        %
                    % \right.
                \end{align}

                  % \captionof{figure}{Another figure}
                  % \label{fig:test2}
            \end{minipage}
        \end{framed}
        \captionof{figure}{Graphical view and summary of the \LBM extended to MNAR missingness process. The observed data $\NXo_{\Ni\Nj}$ is generated by the necessary information carried by the class and propensity of row $\Ni$ and by the class and propensity of the column $\Nj$.}
        \label{fig:graphiclbm}
        \end{figure}
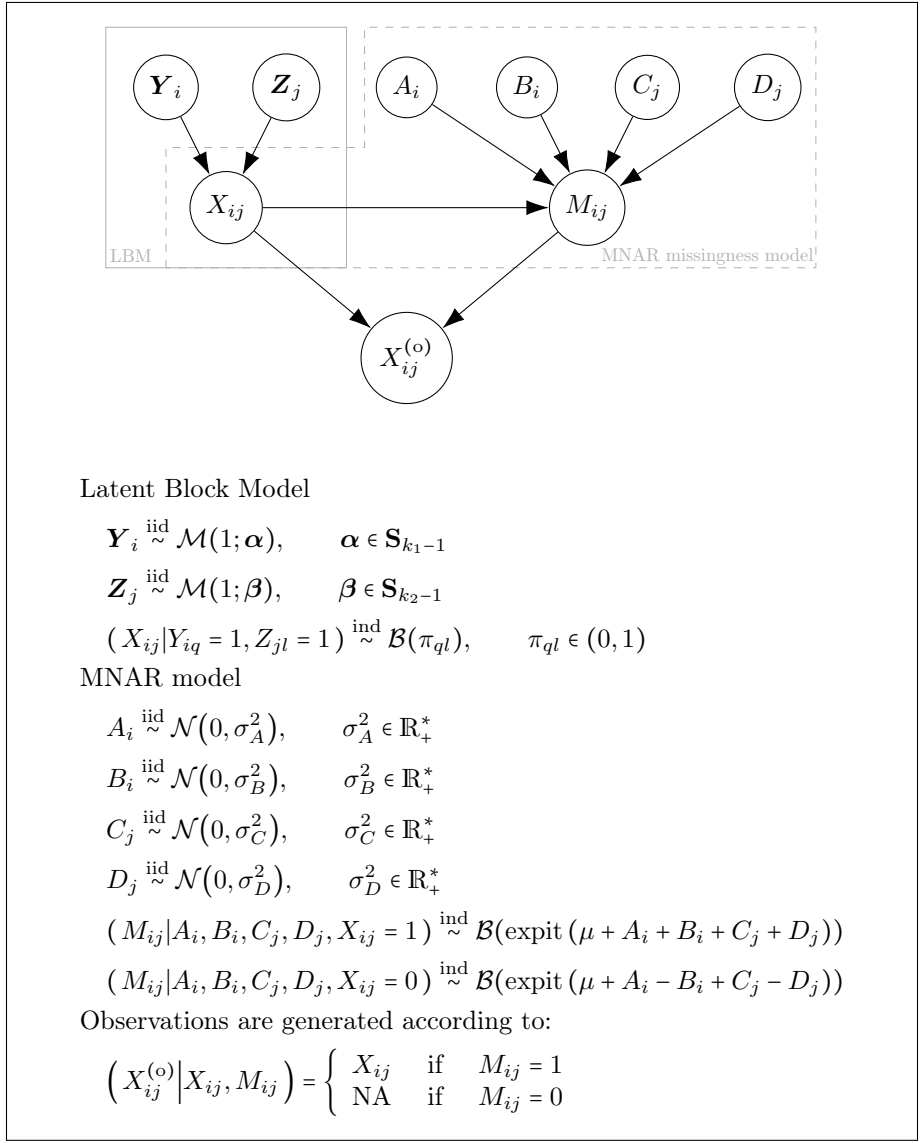

        Given the full matrix $\NXcb$ and the mask matrix $\NMb$, all the elements of the observed matrix $\NXob$ are independent and identically distributed:

        \begin{equation}
            %\small
            \p{\sachant{ \NXo_{\Ni\Nj} }{ \NXc_{\Ni\Nj}, M_{\Ni\Nj}}}
            =
            \left\{
              \begin{array}{cll}
                \NXc_{\Ni\Nj} &\text{ if } \quad \NM_{\Ni\Nj} = 1 \\
                \NNA  &\text{ if } \quad \NM_{\Ni\Nj} = 0 \\
              \end{array}
            \right.
            \enspace.
        \end{equation}
        Figure~\ref{fig:graphiclbm} summarizes the \LBMacro extented to MNAR data.

        $\NXob$ taking its values in $(0,1,\NNA)$, the same model can be rewritten with a Categorial distribution using directly the latent variables of both models:
    
        {%\small
        \begin{align}
                % \left\{
                    \forall \Ni,\;\NYoneb_\Ni &\iidsim \multinomial1{\Nalphaoneb} \nonumber\\
                    \forall \Nj,\; \NYtwob_\Nj &\iidsim \multinomial1{\Nalphatwob}\nonumber\\
                    \forall \Ni,\;\NA_\Ni &\iidsim \norm0{\NsigmaA} \nonumber\\ \label{eq:reform}
                    \forall \Ni,\;\NB_\Ni &\iidsim \norm0{\NsigmaB}  \\
                    \forall \Nj,\;\NP_\Nj &\iidsim \norm0{\NsigmaP} \nonumber\\
                    \forall \Nj,\;\NQ_\Nj &\iidsim \norm0{\NsigmaQ} \nonumber\\
                    \forall \Ni,\Nj, \;
                    \sachant{ \NXo_{\Ni\Nj} }{\NYone_{\Ni\Nq}=1,\NYtwo_{\Nj\Nl}=1,\NA_\Ni,\NB_\Ni,\NP_\Nj,\NQ_\Nj  }
                    &\indsim
                    \categorial{
                    \begin{bmatrix}0\\1\\\NNA\end{bmatrix}
                        }{
                        \begin{bmatrix}
                            p_0 \\
                            p_1 \\
                            1- p_0 - p_1
                        \end{bmatrix}
                        }\nonumber
                % \right. \\
        \end{align}}
        with
        {%\small
        \begin{align}
            p_0 &= \p{1-\Npi_{\Nq\Nl}}\logistic\p{\Nmu + \NA_\Ni- \NB_\Ni +\NP_\Nj - \NQ_\Nj} \label{eq:p2} \\
            p_1 &= \Npi_{\Nq\Nl}\logistic\p{\Nmu + \NA_\Ni+ \NB_\Ni +\NP_\Nj + \NQ_\Nj} \label{eq:p1}
            \enspace.
        \end{align}}

\section{Inference in the extented \LBMacro}
    \label{sec:inference}
    The dependency between the full data matrix $\NXcb$ and the mask matrix $\NMb$  requires a joint inference of the \LBMacro with the \NMARacro model.
    As the standard maximum likelihood approach cannot be applied directly, we adopt a strategy based on a variational EM. % with additional approximations.
    \bigskip
    
    %During inference, we use the reformulation (Equation~\ref{eq:reform}) of the joint model for computational reasons.
    During inference, we use the reformulation (Equation~\ref{eq:reform}).
    We can split our random variables into two sets: the set of unobserved latent variables and the set of observed variables consisting of $\NXob$ only.
    An observation of $\NXob$ only is called the incomplete data, and an observation of $\NXob$ together with the latent variables $\NAb$, $\NBb$, $\NPb$, $\NQb$, $\NYoneb$ and $\NYtwob$ is called the complete data.
    Given the incomplete data, our objective is to infer the model parameters $\theta$ via maximum likelihood {$\Nthetahat=\argmax_\Ntheta p(\NXob;\Ntheta)$}.
    %However, the log-likelihood is not tractable as it involves an exponentially growing sum over all possible values of the latent variables.
    %The likelihood of this incomplete data set can be obtain by summing $p(\NXob | \NAb, \NBb, \NPb, \NQb, \NYoneb, \NYtwob; \Ntheta)$ over all possible %values of the latent variables.
    %As this summation is intractable, we use the EM algorithm to maximize $p(\NXob; \Ntheta)$ without explicitely calculating it.
    We resort to the Expectation Maximization (EM) algorithm to maximize $p(\NXob; \Ntheta)$ without explicitely calculating it.
    The EM algorithm iteratively applies the two following steps:
    \begin{description}[align=left]
        \item[E-step] Expectation step: from the current estimate $\Ntheta^{(t)}$ of $\theta$, compute the criterion  {$\mathcal{Q}(\Ntheta|\Ntheta^{(t)})$} defined as the expectation of the complete log-likelihood, conditionally on the observations $\NXob$:
        \[\mathcal{Q}(\Ntheta|\Ntheta^{(t)})=\expectation_{\sachant{\NYoneb, \NYtwob, \NAb, \NBb, \NPb, \NQb}{\NXob, \Ntheta^{(t)}}}\brackets{\llikli\p{\NXob, \NYoneb, \NYtwob, \NAb, \NBb, \NPb, \NQb; \Ntheta}}\]
        \item[M-step] Maximization step: find the parameters that maximize {$\mathcal{Q}(\Ntheta|\Ntheta^{(t)})$}. \\
                \[\theta^{(t+1)}=\argmax_{\Ntheta}\;\mathcal{Q}(\Ntheta|\Ntheta^{(t)})\]
    \end{description}

    The computation of the complete log-likelihood at the E-step requires the posterior distribution of the latent variables $p(\NYoneb, \NYtwob, \NAb, \NBb, \NPb, \NQb|\NXob)$ which is intractable, because the search space of the latent variables is combinatorially too large.
    %The computation of the complete log-likelihood at the E-step requires the posterior distribution of the latent variables: 
    %\begin{equation}
        %\small
     %   p(\NYoneb, \NYtwob, \NAb, \NBb, \NPb, \NQb|\NXob) = 
     %   \frac{p(\NXob|\NYoneb, \NYtwob, \NAb, \NBb, \NPb, \NQb)p(\NYoneb, \NYtwob, \NAb, \NBb, \NPb, \NQb)}{p(\NXob)}
    %\end{equation}
    %with {\small$p(\NXob) = \sum_{\NYoneb \NYtwob} \int_{\NAb\;\NBb\;\NPb\;\NQb} p(\NXob, \NYoneb, \NYtwob, \NAb, \NBb, \NPb, \NQb)$}.
    %The marginalization over {\small$\NYoneb$, $\NYtwob$, $\NAb$, $\NBb$, $\NPb$, $\NQb$} to calculate {\small$p(\NXob)$} in the denominator is intractable, because the search space of the latent variables is combinatorially too large.
    %So in practice an exact calculation of the E-step is prohibitively expensive as the dimensionality of the latent space is too high to work with directly.
    This problem is well known in the context of co-clustering; for the \LBM, 
    \cite{celeuxd85, keribin:hal-00802764} propose a stochastic E-step with Monte Carlo sampling, but this strategy is not suited to  large-scale problems.
    We follow the original strategy proposed by \citet{lbmgg}, which relies on a variational formulation of the problem, 
    %where a restriction on the explored space to provide an analytical approximation to the posterior distribution of the latent variables. We resort to a variational inference as 
    since it is more efficient in high dimension.

    \subsection{Variational EM}
    The variational EM (VEM) \citep{Jordan,Jaakkola00tutorialon} introduces $q(\cdot)$, a parametric inference distribution defined over the latent variables $\NYoneb$, $\NYtwob$, $\NAb$, $\NBb$, $\NPb$, $\NQb$ and optimize the following lower bound on the log-likelihood of the incomplete data:
    \begin{equation}
        %\small
        \mathcal{J}\p{q, \Ntheta} = \llikli\p{\NXob; \Ntheta} - KL\p{q(\cdot) \parallel p(\cdot | \NXob; \Ntheta)}
        \enspace,
    \end{equation}
    where $KL$ stands for the Kullback-Leibler divergence and $q(\cdot)$ denotes the variational distribution over the latent variables $\NYoneb$, $\NYtwob$, $\NAb$, $\NBb$, $\NPb$ and $\NQb$.
    It can be shown that $\mathcal{J}\p{q, \Ntheta}$ is a concave function of the variational distribution $q$ and that its maximum is reached for $q(\cdot) = p(\cdot | \NXob; \Ntheta)$.
    Thus, maximizing the criterion $\mathcal{J}$ is equivalent to minimizing the discrepancy between $q(\cdot)$ and $p(\cdot | \NXob; \Ntheta)$, as measured by the Kullback divergence, and is also equivalent to maximizing the likelihood.
    The minimization of this Kullback divergence requires to explore the whole space of latent distributions; the difficulty of the problem is equivalent, in terms of complexity, to the initial problem.

    The criterion $\mathcal{J}\p{q, \Ntheta}$ can also be expressed as the sum of a negative \guill{energy} and the entropy of $q$ hence its name \guill{negative variational free energy} in analogy with the thermodynamic free energy:
    \begin{equation}
        %\small
        \label{eq:criterionj}
        \mathcal{J}\p{q, \Ntheta} = \mathcal{H}(q) + \expectation_q \brackets{\llikli\p{\NXob, \NYoneb, \NYtwob, \NAb, \NBb, \NPb, \NQb; \Ntheta}}
        \enspace,
    \end{equation}
    where $\mathcal{H}(q)$ is the entropy of the variational distribution and $ \expectation_q$ is the expectation with respect to the variational distribution.
    % If we decide to restrict the space of the latent distributions $q$ to be explored, the criteria can become tractable.
    The criteria $\mathcal{J}$ can become tractable if an exploration of a subspace, noted $\NRX$, of the latent distributions is made.
    However, this solution comes with the cost that the maximum found, will be a lower bound of the initial criteria:
    \begin{equation}
        %\small
        \mathcal{J}\p{q, \Ntheta} \geq \mathcal{J}\p{\NRX, \Ntheta}
    \end{equation}
    $\mathcal{J}\p{\NRX, \Ntheta}$ is also known as the \guill{Evidence Lower BOund} (ELBO) emphasizing  the lower bound property on the evidence of the data.
    %In practice, if the variational space is properly reduced, the method gives estimates that are competitive with Monte Carlo sampling at greater speed. \todo{TROUVER UNE REF ??}

    A wise choice of the restriction on the variational distribution leads a feasible computation of the criterion. We choose to consider the following posterior shapes on the latent variables:
    {%\small
      \begin{align*}
        \forall&\Ni& \NYoneb_\Ni|\NXob &\underset{\NRX}\sim\multinomial1{\Ntauone_\Ni} \\
        \forall&\Nj& \NYtwob_\Nj|\NXob &\underset{\NRX}\sim\multinomial1{\Ntauone_\Nj} \\
        \forall&\Ni& \NA_\Ni|\NXob &\underset{\NRX}\sim\norm{\NnuA_\Ni}{\NrhoA_\Ni} \\
        \forall&\Ni& \NB_\Ni|\NXob &\underset{\NRX}\sim\norm{\NnuB_\Ni}{\NrhoB_\Ni} \\
        \forall&\Nj& \NP_\Nj|\NXob &\underset{\NRX}\sim\norm{\NnuP_\Nj}{\NrhoP_\Nj} \\
        \forall&\Nj& \NQ_\Nj|\NXob &\underset{\NRX}\sim\norm{\NnuQ_\Nj}{\NrhoQ_\Nj} \enspace.
    \end{align*}}
    We also impose the conditional independence of the latent variables to get a feasible computation of the entropy and of the negative “energy” (Equation~\ref{eq:criterionj}) under $\NRX$.
    This conditional independence is widely known as the \guill{mean field approximation} \citep{Parisi}.
    We finally get the following fully factorized shape:
    {%\small
      \begin{align*}
        \NRX &=\textstyle
        \prod_{\Ni=1}^{\Nnone}{\multinomial{1}{\Ntauone_\Ni}}\;\times \;\;
        \prod_{\Nj=1}^{\Nntwo}{\multinomial{1}{\Ntautwo_\Nj}} \\
        &\textstyle\quad\times
        \prod_{\Ni=1}^{\Nnone}{\norm{\NnuA_\Ni}{\NrhoA_\Ni}}\times
        \prod_{\Ni=1}^{\Nnone}{\norm{\NnuB_\Ni}{\NrhoB_\Ni}} \nonumber\\ 
        &\textstyle\quad \times
        \prod_{\Nj=1}^{\Nntwo}{\norm{\NnuP_\Nj}{\NrhoP_\Nj}}\times
        \prod_{\Nj=1}^{\Nntwo}{\norm{\NnuQ_\Nj}{\NrhoQ_\Nj}} \nonumber
        \enspace,
    \end{align*}}
    \noindent
    where {%\small
    $\gamma = (\Ntauoneb, \Ntautwob,\NnuAb,\NrhoAb, \NnuBb, \NrhoBb, \NnuPb, \NrhoPb, \NnuQb, \NrhoQb )$} denotes the parameters concatenation of the restricted variational distribution $\NRX$.
    
    %to a feasible computation of the entropy $\mathcal{H}(q)$ and a feasible computation of the expectation under the variational distribution.
    %Restricting the variational distribution to have a fully factorized shape is appropriate as the entropy is additive across independent variables.
    %It is commonly called the \textit{mean field approximation}.
    %The variational distribution $q$ of our model is restricted to a class of distributions $\NRX$ having the following shape:
    %\begin{align}
    %        \NRX
    %        &=\prod_{\Ni=1}^{\Nnone}{\multinomial{1}{\Ntauone_\Ni}}
    %        \prod_{\Nj=1}^{\Nntwo}{\multinomial{1}{\Ntautwo_\Nj}} \nonumber\\
    %        & \times
    %        \prod_{\Ni=1}^{\Nnone}{\norm{\NnuA_\Ni}{\NrhoA_\Ni}}
    %        \prod_{\Ni=1}^{\Nnone}{\norm{\NnuB_\Ni}{\NrhoB_\Ni}}  \nonumber\\
    %        & \times
    %        \prod_{\Nj=1}^{\Nntwo}{\norm{\NnuP_\Nj}{\NrhoP_\Nj}}
    %        \prod_{\Nj=1}^{\Nntwo}{\norm{\NnuQ_\Nj}{\NrhoQ_\Nj}}
    %\end{align}
    %The posteriors of the latent variables $\NAb$, $\NBb$, $\NPb$ and $\NQb$ are approximated by Gaussian distributions and the posteriors of $\NYoneb$ and $\NYtwob$ by multinomial distributions.
    %We denote by $\gamma$ the vector ($\Ntauone$, $\Ntautwo$, $\NnuA$, $\NrhoA$, $\NnuB$, $\NrhoB$, $\NnuP$, $\NrhoP$, $\NnuQ$, $\NrhoQ$) of parameters of the distribution $\NRX$.
    \bigskip

    The new criteria $\mathcal{J}\p{\gamma, \Ntheta}$ that we want to optimize from now on is:
    \begin{equation}
        %\small
        \mathcal{J}\p{\gamma, \Ntheta} = \mathcal{H}(\NRX) + \expectation_{\NRX}\left[\llikli\p{\NXob, \NYoneb, \NYtwob, \NAb, \NBb, \NPb, \NQb; \Ntheta}\right]
        \label{eq:criteriaJ}
    \end{equation}
    and the initial estimates of the model parameters $\Nthetahat$ are inferred as:
    \begin{equation}
        %\small
        \Nthetahat = \underset{\theta}{\argmax\;} \p{\underset{\gamma}{\max\;} \mathcal{J}\p{\gamma, \Ntheta}}
        \enspace.
    \end{equation}
    This double maximization is realized with an iterative strategy and can be seen as an extension of the EM algorithm. The two steps are described in Algorithm~\ref{algo:vem}.

    \begin{algorithm}[tb]
    \SetAlgoLined
    \textbf{Input}: observed data $\NXob$, $\Nnq$ and $\Nnl$ number of row groups and column groups \;
    - Initialize $\gamma^{(0)}$ and $\Ntheta^{(0)}$\;
    - \While{not convergence of criterion $\mathcal{J}$}{
        VE-step: find the variational parameters $\gamma^{(t+1)}$ that optimize $\mathcal{J}\p{\gamma, \Ntheta^{(t)}}$
            \[\gamma^{(t+1)} = \underset{\gamma}{\argmax\;} \mathcal{J}\p{\gamma, \Ntheta^{(t)}}\]
        M-step: find the model parameters $\Ntheta^{(t+1)}$ that optimize $\mathcal{J}\p{\gamma^{(t)}, \Ntheta}$:
            \[ \Ntheta^{(t+1)} = \underset{\Ntheta}{\argmax\;} \mathcal{J}\p{\gamma^{(t)}, \Ntheta}\]
     }
     \KwResult{$\theta$ and $\gamma$: model and variational parameters}
     \caption{Variational Expectation Maximization algorithm.}
     \label{algo:vem}
    \end{algorithm}

    \subsection{Computation of the variational criterion}
    The restriction on the space of the variational distribution simplifies the computation of $\mathcal{H}(\NRX)$ as entropy is additive across independent variables:
     \begin{multline*}
      \mathcal{H}(\NRX) =
            - \sum_{\Ni\Nq}{ \Ntauone_{\Ni\Nq} \log\Ntauone_{\Ni\Nq}}
            - \sum_{\Nj\Nl}{ \Ntautwo_{\Nj\Nl} \log\Ntautwo_{\Nj\Nl}}
            + \frac{1}{2} \sum_{\Ni}{\log\p{2\pi e \NrhoA_\Ni }} \\
            + \frac{1}{2} \sum_{\Ni}{\log\p{2\pi e \NrhoB_\Ni }}
            + \frac{1}{2} \sum_{\Nj}{\log\p{2\pi e \NrhoP_\Nj }} + \frac{1}{2} \sum_{\Nj}{\log\p{2\pi e \NrhoQ_\Nj }} \enspace.
     \end{multline*}
    The independence of latent variables allows to rewrite the expectation of the complete log-likelihood as:
     \begin{multline}\label{equ:expectationall}
     \expectation_{\NRX}\brackets{\llikli\p{\NXob, \NYoneb, \NYtwob, \NAb, \NBb, \NPb, \NQb}} =
            \;\expectation_{\NRX}\brackets{\llikli\p{\NYoneb}}\\ 
            + \expectation_{\NRX}\brackets{\llikli\p{\NYtwob}}
            +\expectation_{\NRX}\brackets{\llikli\p{\NAb}}
            + \expectation_{\NRX}\brackets{\llikli\p{\NBb}} \\
            + \expectation_{\NRX}\brackets{\llikli\p{\NPb}}
            + \expectation_{\NRX}\brackets{\llikli\p{\NQb}}
            +\expectation_{\NRX}\brackets{\llikli\p{ \sachant{\NXob}{\NYoneb, \NYtwob, \NAb, \NBb, \NPb, \NQb}}  }
            \enspace.
     \end{multline}
     
    Despite the variational approximation, the expectation of the complete log-likelihood~\eqref{equ:expectationall} can not be exactly computed as its last term involves an expectation under $\NRX$ of nonlinear functions:
     \begin{multline}\label{equ:expectation_cond}
        \expectation_{\NRX}\brackets{\llikli\p{ \sachant{\NXob}{\NYoneb, \NYtwob, \NAb, \NBb, \NPb, \NQb}}} = 
        \hspace{-10pt} \sum_{\Ni\Nj\Nq\Nl:\NXo_{\Ni\Nj}=0} \hspace{-10pt} \Ntauone_{\Ni\Nq}\Ntautwo_{\Nj\Nl} \expectation_{\NRX}\brackets{\log\p{ p_0}} \\
        + \hspace{-10pt} \sum_{\Ni\Nj\Nq\Nl:\NXo_{\Ni\Nj}=1} \hspace{-10pt}  \Ntauone_{\Ni\Nq}\Ntautwo_{\Nj\Nl} \expectation_{\NRX}\brackets{\log\p{p_1}}
        + \hspace{-10pt} \sum_{\Ni\Nj\Nq\Nl:\NXo_{\Ni\Nj}=\NNA}  \hspace{-10pt} \Ntauone_{\Ni\Nq}\Ntautwo_{\Nj\Nl}  \expectation_{\NRX}\brackets{\log\p{ 1 - p_0 -p_1 }}
    	\enspace,
     \end{multline}
     with $p_0$ and $p_1$ defined in Equations \eqref{eq:p2}--\eqref{eq:p1}.

    These expectations can be approximated by the delta method \cite[p. 79]{deltamethod}. Using a first order Taylor expansion would lead to a criterion without maximum, so we use a second order Taylor expansion.
%\iffalse
%Applying the Delta method with second order Taylor developments on any function $f$ with $X$ and $Y$ independent random variables, leads to:
%    \begin{small}
%        \begin{equation}
%            \expectation_{\NRX}\brackets{f\p{X,\; Y}}  \approx f\p{\expectation X,\; \expectation Y}
%            + \frac{1}{2} \; var(X) \; \frac{\partial^2  f\p{\expectation X,\;\expectation Y}}{\partial \p{X}^2}
 %           + \frac{1}{2} \; var(Y) \; \frac{\partial^2  f\p{\expectation X,\; \expectation Y}}{\partial \p{Y}^2}
%            \label{eq:delta}
%        \end{equation}
%    \end{small}
%    Using Equation~\eqref{eq:delta}, the Delta method is applied to compute the expectations of the functions~\eqref{eq:f1}, \eqref{eq:f0} and \eqref{eq:fna}.
%    With these approximations, the optimization criterion $\mathcal{J}\p{\gamma, \Ntheta}$ is fully tractable.
%\fi
    The full expression of the criterion is given in Appendix~ \ref{annexcriteria}.

    \subsection{Maximization of the variational criterion}

    The VEM Algorithm~\ref{algo:vem} alternates between a maximization with respect to the variational parameters $\Ngamma$ and a maximization w.r.t the model parameters $\Ntheta$.
    % The VEM algorithm is an alternated double maximization: one w.r.t the variational parameters $\Ngamma$ and one w.r.t the model parameters $\Ntheta$.
    For our model, there is no explicit solution for the two maximizations of the criterion $\mathcal{J}\p{\gamma, \Ntheta}$, which are are carried out by the Limited-memory Broyden–Fletcher–Goldfarb–Shanno (L-BFGS) algorithm.
    %We used the Limited-memory Broyden–Fletcher–Goldfarb–Shanno (L-BFGS) algorithm which is a quasi-Newton method that realizes a numerical maximization using a limited amount of computer memory as it replaces the Hessian matrix computation with an estimate \citep{lbfgs}.
    We used automatic differentiation to compute the gradients needed for L-BFGS and for the Taylor series used in the variational criterion.
    We chose the Autograd library from HIPS and the submodule Autograd from PyTorch \citep{NEURIPS2019}.
    These libraries rely on a reverse accumulation computational graph to compute exact gradients.
    Their high efficiency, even with large graphs, thanks to GPU acceleration, makes them particularly well adapted for the VEM algorithm.

    \subsection{Initialization} \label{initproc}

    VEM does not ensure convergence towards a global optimum.
    The EM-like algorithms are known to be sensitive to the initialization, particularly when applied to models with discrete latent space, and may get stuck into unsatisfactory local maxima \citep{Biernacki2003,Baudry2015}. 
        
    A simple solution consists in 
    training for a few iterations from several random initializations, and pursue optimization with the solution with highest value of the variational criterion \citep[see, e.g., small EM for mixtures][]{Baudry2015}.
    This exploration strategy spends a great deal of computing resources to bring out only a few good estimates. 
    Another solution is to rely on simpler clustering methods, such as k-means or spectral clustering, to initialize the algorithm \citep{Shireman}.
    
    For the Stochastic Block Model, a close relative of the Latent Block Model for graphs, \citet{Rohe_2011} prove the consistency of spectral clustering to identify the parameters of the Stochastic Block Model. Following this idea, we use a double spectral clustering (with absolute eigenvalues of the Laplacian as \citet{Rohe_2011}) on rows and columns on similarity matrices, to initialize our algorithm.
    Although this method is not designed for \NMARacro data, it can be expected to provide a satisfying initialization of the \LBM if the missingness is not predominant. 
    The parameters of our missingness model can not be initialized with this procedure; they are randomly initialized.
    The overall initialization procedure is described in Appendix~ \ref{annex:init}.
    \section{Model selection}
    \label{sec:modelselection}
    \subsection{Integrated Completed Likelihood criterion (ICL) }
    \label{sec:criterion}
    ICL, inspired by the Bayesian Information Criterion, 
    % approximates the complete data integrated log-likelihood 
    was originally proposed to select a relevant number of classes for mixture models \citep{ICLbiernacki}.
    It was extended to select an appropriate number of (row and column) clusters in the standard \LBM  \citep{keribinicl}:
    for $\Nnq$ row classes and $\Nnl$ column classes, the criterion reads
    {%\small
    \begin{align}
        ICL(\Nnq, \Nnl) & = \llikli\p{\NXb, \NYoneb, \NYtwob} \nonumber \\
        & = \log \int \Prob{\sachant{\NXb, \NYoneb, \NYtwob}{\Ntheta; \Nnq, \Nnl}} \Prob{\Ntheta; \Nnq, \Nnl}\mathrm{d}\Ntheta
        \enspace,
    \end{align}}
    with $\Prob{\Ntheta; \Nnq, \Nnl}$ the prior distribution of parameters.
    By taking into account the latent variables $\NYoneb, \NYtwob$, ICL is a clustering-oriented criterion , whereas  BIC or AIC are driven by the faithfulness to the distribution of $\NXb$ \citep{ICLbiernacki}.

    For the \LBMacro with \NMARacro missingness, ICL requires priors on the parameters of the missingness model. We chose independent and non informative InverseGamma($\beta$, $\beta$) distribution (where $\beta$ tends to zero) for the parameters $\NsigmaA$, $\NsigmaB$, $\NsigmaP$ and $\NsigmaQ$. 
    As in \citep{keribinicl}, we use non-informative Dirichlet distribution priors on the parameters $\Nalphaoneb$ and $\Nalphatwob$ of mixing proportions of classes.
    ICL reads 
    \begin{align}
        ICL(\Nnq, \Nnl) &= \llikli\p{\NXb, \NYoneb, \NYtwob, \NAb, \NBb, \NPb, \NQb} \\  
        &= \log \int \Prob{\sachant{\NXb, \NYoneb, \NYtwob, \NAb, \NBb, \NPb, \NQb}{\Ntheta; \Nnq, \Nnl}} \Prob{\Ntheta; \Nnq, \Nnl}\mathrm{d}\Ntheta \nonumber
    \end{align}

    \begin{proposition}
        The ICL criterion for  the \LBMacro extended to the \NMARacro missingness process presented in Section~\ref{sect:lbmextended} has the following asymptotic 
        form for a data matrix of size $\Nnone \times \Nntwo$:
        {%\small
        \begin{align*}
        ICL(\Nnq, \Nnl) =&\;
        \underset{\Ntheta}{\max}\; \log \Prob{\NXob,\NYoneb, \NYtwob, \NAb, \NBb, \NPb, \NQb; \Ntheta} \nonumber\\
        & - \frac{\Nnq\Nnl}{2} \log\p{\Nnone\Nntwo} %\nonumber\\
         - \frac{\Nnq-1}{2} \log\p{\Nnone} - \frac{\Nnl-1}{2} \log\p{\Nntwo} \label{equ:iclAsympt}\\
        & + \Nnone \log\p{2\pi} - \log\p{\Nnone} %\nonumber\\
        + \Nntwo \log\p{2\pi} - \log\p{\Nntwo} \nonumber\\
        & + o(\log\Nnone) + o(\log\Nntwo) \enspace.
    \end{align*}}
    See proof in Appendix~\ref{annexicl}.
    \end{proposition}

    Since the maximum of the complete log-likelihood required to calculate the ICL is not available, in practice it is replaced by the lower bound provided by the variational approximation (see equation~\ref{eq:criteriaJ}). An ICL criterion for the \LBMacro with MAR missing data can be constructed in the same way, allowing for comparison with the \NMARacro model (see details in Appendix~\ref{annexicl}).

\section{Experiments on simulated data}
\label{sec:simulatedata}
    Simulated data brings all the elements to assess clustering algorithms in controlled settings.
    Using controlled datasets provides the means to properly test the ability of an algorithm to recover the known underlying structure. 
    \subsection{Assessing the difficulty of a co-clustering task}
In co-clustering, several loss functions are suited for measuring the discrepancy between the underlying classes ($\NYoneb$, $\NYtwob$) and some predictions ($\NYonebtild$, $\NYtwobtild$).  For our experiments, we will use the measure defined by \cite{lbmgg}, that is, the ratio of misclassified entries in the data matrix:
    \begin{equation*}
            \Nlossitem{\NYoneb, \NYtwob,\NYonebtild, \NYtwobtild}
            =
            \overbrace{\Nlossrow{\NYoneb, \NYonebtild}}^{1 - \frac{1}{\Nnone} \sum_\Ni\Kronecker{\NYone_\Ni, \NYonetild_\Ni} }
            +\overbrace{\Nlosscol{\NYtwob, \NYtwobtild}}^{1 - \frac{1}{\Nntwo} \sum_\Nj\Kronecker{\NYtwo_\Nj, \NYtwotild_\Nj} } 
            -\;\Nlossrow{\NYoneb, \NYonebtild} \;\Nlosscol{\NYtwob, \NYtwobtild} 
    \end{equation*}
where $\Kronecker{}$ is the Kronecker delta.

In standard clustering, the difficulty of a task is often assessed by its Bayes risk, that is, by the minimum of the expectation of the loss function, which is typically approximated by Monte Carlo on simulated data. Co-clustering poses specific difficulties.
%A Bayes classifier returns the labels $\NYonebtild$, $\NYtwobtild$ that minimize its associated loss function. In order to assess the difficulty of a clustering task, the Bayes risk, that is the expected loss for the Bayes classifier is often used. 
    %In order to assess the difficulty of a clustering task, the Bayes risk is often used.
    %The Bayes risk is the expected loss for the Bayes classifier.
    %
    %Several loss functions may be defined to measure the dissimilarity between the true labels ($\NYoneb$, $\NYtwob$) and the  labels predicted by the Bayes classifier ($\NYonebtild$, $\NYtwobtild$). For co-clustering, several definitions may be given;
    %we will use for our experiments the simple discrepancy measure defined by \cite{lbmgg}, that is, the ratio of misclassified entries in the data matrix:
    % such as the Rand Index or
    %\begin{equation*}
    %        \Nlossitem{\NYoneb, \NYtwob,\NYonebtild, \NYtwobtild}
    %        =
    %        \overbrace{\Nlossrow{\NYoneb, \NYonebtild}}^{1 - \frac{1}{\Nnone} \sum_\Ni\Kronecker{\NYone_\Ni, \NYonetild_\Ni} }
    %        +\overbrace{\Nlosscol{\NYtwob, \NYtwobtild}}^{1 - \frac{1}{\Nntwo} \sum_\Nj\Kronecker{\NYtwo_\Nj, \NYtwotild_\Nj} } 
    %        -\;\Nlossrow{\NYoneb, \NYonebtild} \;\Nlosscol{\NYtwob, \NYtwobtild} 
    %\end{equation*}
    %where $\Kronecker{}$ is the Kronecker delta.
    %
    %Assessing the difficulty of a co-clustering task from a generated data set is not obvious.
    Adding more rows or more columns alter its difficulty because the dimensions of the spaces where the clustering is performed are expanded.
    The duality between the rows and the columns imply that the size of the matrix is a characteristic of a co-clustering problem.
    In other words, given a fixed generative distribution, as the matrix size increases, the difficulty of the task decreases, in contrast to simple clustering, where the difficulty, as measured by the Bayes risk, remains constant when more examples (that is, rows) are added.    
    
    A simple Monte Carlo approximation of the risk consists in averaging over many statistical units.
    In simple clustering, this means generating a great number of rows in a data matrix.
    In co-clustering, the statistical unit is the whole matrix, implying that a Monte Carlo approximation of the risk is obtained by generating a great number of data matrices; which then involves a great computational time.
    Furthermore, estimating the Bayes risk from a single data matrix is very inconstant; the risk may be very different between two data matrices of same size generated from the same distribution.
    Hence the usual notion of Bayes risk is not appropriate for co-clustering.
    \citet{Lomet12} argue that conditioning the Bayes risk on the observed matrix is more appropriate.
    They give a protocol to simulate data matrices in which the difficulty of the clustering task is controlled by the following {\em conditional Bayes risk}:
    \begin{equation}\label{eq:condBayesrisk}
        r_{item}(\NYonebtild, \NYtwobtild) = \expectation\brackets{
                                                                \sachant{\Nlossitem{
                                                                \NYoneb, \NYtwob, \NYonebtild, \NYtwobtild
                                                                }}
                                                                {\NXob}}
                                                                \enspace,
    \end{equation}
    where the expectation is taken over $\NYoneb, \NYtwob$ only and $\NYonebtild,\NYtwobtild$ are the  clusterings returned by the {\em conditional Bayes classifier}, that is, the maximum \textit{a posteriori}:
    \begin{equation*}
    \p{\NYonebtild,\NYtwobtild} = \argmin_\p{\NYoneb, \NYtwob} r_{item}(\NYoneb, \NYtwob) = \argmax_\p{\NYoneb, \NYtwob} \sum_{\Ni\Nj} p\p{\sachant{\NYone_\Ni, \NYtwo_\Nj}{\NXob}}
        \enspace.
    \end{equation*}

    \citet{Lomet12} released data sets, with different sizes and difficulties, simulated from the \LBM.
    Using their protocol, we generated new data according the \LBMacro with a \NMARacro missingness process.
    Data sets are generated according to the \LBMacro with three row and column classes, with parameters
    \begin{equation}\label{eq:LBMparam}
        \Nalphaoneb = \Nalphatwob =
                \begin{pmatrix}
                        \sfrac13 \\
                        \sfrac13 \\
                        \sfrac13
                \end{pmatrix}
        \qquad\text{and}\qquad
            \Npib =
            \begin{pmatrix}
                \epsilon & \epsilon & 1-\epsilon \\
                \epsilon & 1-\epsilon & 1-\epsilon \\
                1-\epsilon & 1-\epsilon & \epsilon \\
            \end{pmatrix}
            \enspace,
    \end{equation}
    where $\epsilon$ defines the difficulty of the clustering task.
    The parameters of the \NMARacro process are
    \begin{equation}\label{eq:MNARparam} 
        \Nmu  = 1 , \quad \NsigmaA  = 1 , \quad \NsigmaB  = 1 , \quad \NsigmaP  = 1 , \quad \NsigmaQ = 1 \enspace,
    \end{equation}
    which gives an average proportion of 35\% of missing values.
    \subsection{Analyzing the classification of the proposed inference}
    \label{sec:classifinference}
\begin{figure}%[h]
    \centering
        \scalebox{0.5}{\input{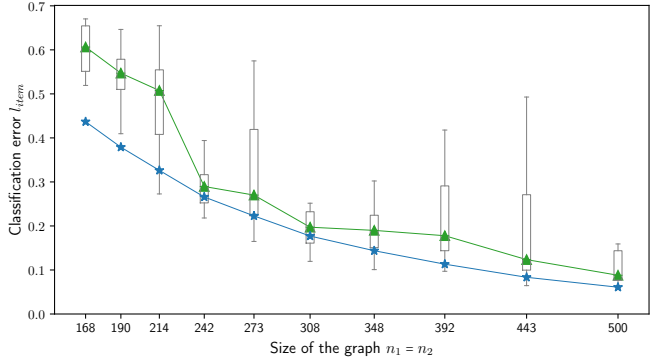}}
        \caption{Classification error with respect to the size of the data matrix (lower is better);
        \textcolor[rgb]{0.121569,0.466667,0.705882}{$\filledstar$} is the median of the conditional Bayes risk;
        \textcolor[rgb]{0.172549,0.627451,0.172549}{$\filledmedtriangleup$} is the median prediction error obtained by our algorithm.}
		\label{fig:error_classif_size}
    \end{figure}

    We test here the ability of the proposed inference scheme to recover row and column classes.
    To conduct the experiments, we generate an initial data matrix of size $\Nnone=\Nntwo=500$ with a conditional Bayes risk of $5\%$ set by choosing $\epsilon$ \eqref{eq:LBMparam} by trial and error.
    The size of this matrix is then progressively reduced, removing rows and columns, to increase the difficulty of the classification task.
    The conditional Bayes risk is re-estimated on each sub matrix to provide a reference.
    Our algorithm is then run on these data matrices using 20 initializations for each run, as described in Section~\ref{initproc}.
    We then predict the row and column classes $(\NYoneb, \NYtwob)$ with their maximum \textit{a posteriori} estimators  on the variational distribution.
    This whole process is repeated 20 times, leading to the results presented in Figure~\ref{fig:error_classif_size}.
    
    As expected, the conditional Bayes risk decreases as the data matrices grow.
    The predictions returned by our algorithm follow the same pattern, with a diminishing gap to the conditional Bayes risk as the data matrices grow, which is consistent with our expectations.
    Appendix~\ref{annex:estimation} provides additional experimental results that show consistent estimations of the model parameters.

    \subsection{Analyzing the benefit of a \NMARacro model versus a MAR model for \NMARacro data} \label{sec:expnmar}
    The importance of using the right missingness model is tested by comparing the classifications returned by an \LBMacro with and without an \NMARacro model.
    A data set is generated according to the \LBMacro with \NMARacro values where the parameters $\Nalphaoneb$, $\Nalphatwob$ and $\Npib$ of the \LBMacro are fixed as in \eqref{eq:LBMparam}, and $\epsilon$ is chosen in order to get a conditional Bayes risk of 12\%, for data matrices of size $\Nnone = \Nntwo = 100$; the \NMARacro model parameters $\Nmu$, $\NsigmaA$ and $\NsigmaP$ all set to one which gives an average proportion of 35\% of missing values.
    Several data matrices are generated using these parameters while varying the value of the $\NsigmaB$ and $\NsigmaQ$ parameters that govern the \NMARacro effects;
    these  variations do not affect the conditional Bayes risk nor the proportion of missing values.
    For each data matrix, we train the \LBMacro with either the \MARacro or the \NMARacro model.
    This process is repeated 20 times, starting from the generation of a new fully observed data matrix.
    
\begin{figure}%[tb]
    \centering
        \scalebox{0.5}{\input{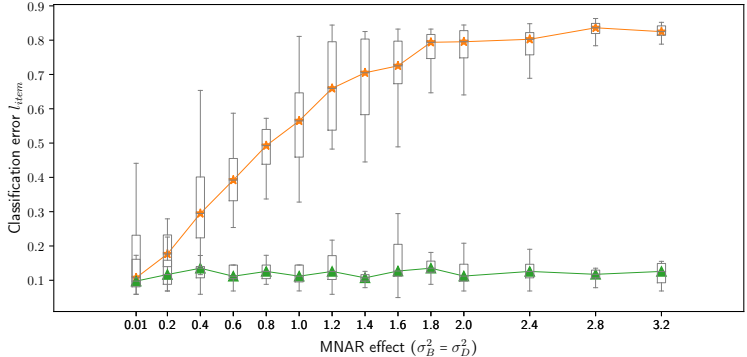}}%
        \caption{Classification error with respect to an increase of the \NMARacro effect (lower is better); \textcolor[rgb]{1.000000,0.498039,0.054902}{$\filledstar$} is the median prediction error obtained with the MAR model; \textcolor[rgb]{0.172549,0.627451,0.172549}{$\filledmedtriangleup$} is the median prediction error obtained with the \NMARacro model.}
		\label{fig:error_classif_nmar}
    \end{figure}

    The median of the classification errors $l_{item}$ are presented in Figure~\ref{fig:error_classif_nmar} as a function of the \NMARacro effect. They are essentially constant and close to the conditional Bayes risk for the \LBMacro with the \NMARacro model, whereas the \LBMacro with the \MARacro model is badly affected by \NMARacro data, eventually leading to a classification close to a totally random allocation \footnote{With equal class proportions, the expected classification error of a random allocation is $\frac{\Nnq-1}{\Nnq} + \frac{\Nnl-1}{\Nnl} - \frac{\Nnq-1}{\Nnq} \frac{\Nnl-1}{\Nnl}$, that is, 0.89 here where $\Nnq=\Nnl=3$.}.
    Ignoring the nature of the missingness process leads here to strong biases in estimation that in turn drastically affect classification. Thankfully, the ICL criterion may be of great help to select the right missingness model as shown in Section~\ref{sec:icl_exp_missing_model}.

    \subsection{Analyzing the ability of the model selection criterion to select the adequate number of classes}
    We reuse the parameters \eqref{eq:LBMparam} and \eqref{eq:MNARparam} to analyze the behavior of the %asymptotic 
    ICL criterion.
    We consider different sizes of data matrices, between (30,30) and (150,150), with varying difficulty for each matrix size, with a conditional Bayes risk \eqref{eq:condBayesrisk} of respectively 5\%, 12\% and 20\%

    The results in Figure~\ref{iclresults} show that, as expected, the ICL criterion tends to select more often the right number of classes as the data matrices get larger and also when classes are more separated.
    We also observe that the ICL criterion tends to be conservative for small data matrices, by underestimating the number of classes.
    It could come to the fact that the size of the matrix is not large enough to consider the asymptotic approximation as valid and/or it could come from the approximations used to compute the log-likelihood $\mathcal{J}$ (variational restriction and delta method).
    
    % Hubert, L., Arabie, P. Comparing partitions. Journal of Classification 
    %
    %
    % The $\epsilon$ parameter is adapted to get some data frames with a progressive clustering difficulty: (+, ++ and +++) which stands for a conditional Bayes risk of respectively 5\%, 12\% and 20\%.
    % For each matrix, the proposed variational algorithm is run with a number of row and column classes varying from two to four.
    \begin{figure}
        \centering
        \setlength{\extrarowheight}{1pt}%
        \scalebox{0.82}{\begin{tabular}{c|c|c|cccc|cccc|cccc|}
            \multicolumn{3}{@{}c}{} &  \multicolumn{4}{@{}c@{}}{$r_{item}(\NYonebtild, \NYtwobtild) =5\%$} & \multicolumn{4}{@{}c@{}}{$r_{item}(\NYonebtild, \NYtwobtild) =12\%$} & \multicolumn{4}{c}{$r_{item}(\NYonebtild, \NYtwobtild) =20\%$} \\[1ex]
            \cline{4-15}
            \multicolumn{3}{c|}{} &  \multicolumn{4}{c|}{$\Nnl$} & \multicolumn{4}{c|}{$\Nnl$} & \multicolumn{4}{c|}{$\Nnl$} \\
            \cline{4-15}
%            ($\Nnone, \Nntwo$)&   &  2&3&4&5 & 2&3&4&5 &2&3&4&5 \\
             \multicolumn{3}{c|}{} & 2&3&4&5 & 2&3&4&5 &2&3&4&5 \\
            \cline{2-15}
            \multirow{4}{*}{$\Nnone=\Nntwo=30$} & \multirow{4}{*}{$\Nnq$}  &2&9&2&&&13&3&&&14&2&&\\
            & & 3 &1& \boxed{7}&&&2& \boxed{2}&&&2& \boxed{1}&&\\
            & & 4 &&&&&&&&&&&&\\
            & & 5 &&&&1&&&&&&&&1\\
            \cline{2-15}
            \multirow{4}{*}{$\Nnone=\Nntwo=40$} & \multirow{4}{*}{$\Nnq$}  &2&4&2&&&17&1&&&17&1&&\\
            & & 3 && \boxed{14}&&&1& \boxed{1}&&&1& \boxed{1}&&\\
            & & 4 &&&&&&&&&&&&\\
            & & 5 &&&&&&&&&&&&\\
            \cline{2-15}
            \multirow{4}{*}{$\Nnone=\Nntwo=50$} & \multirow{4}{*}{$\Nnq$}  &2&1&&&&11&2&&&15&1&2&\\
            & & 3 &2& \boxed{17}&&&& \boxed{7}&&&1& \boxed{1}&&\\
            & & 4 &&&&&&&&&&&&\\
            & & 5 &&&&&&&&&&&&\\
            \cline{2-15}
            \multirow{4}{*}{$\Nnone=\Nntwo=75$} & \multirow{4}{*}{$\Nnq$}  &2&1&&&&9&&&&13&2&&\\
            & & 3 &1& \boxed{17}&&&1& \boxed{9}&&&& \boxed{4}&&\\
            & & 4 &&&&&1&&&&&&&\\
            & & 5 &1&&&&&&&&&&&1\\
            \cline{2-15}
            \multirow{4}{*}{$\Nnone=\Nntwo=100$} & \multirow{4}{*}{$\Nnq$}  &2&&&&&2&&&&11&&&\\
            & & 3 && \boxed{19}&&&& \boxed{18}&&&1& \boxed{7}&1&\\
            & & 4 &&&&&&&&&&&&\\
            & & 5 &&&&1&&&&&&&&\\
            \cline{2-15}
            \multirow{4}{*}{$\Nnone=\Nntwo=150$} & \multirow{4}{*}{$\Nnq$}  &2&1&&&&&1&&&5&1&&\\
            & & 3 && \boxed{14}&2&&1& \boxed{18}&&&1& \boxed{11}&&\\
            & & 4 &&1&&&&&&&&&&\\
            & & 5 &&&&2&&&&&&&&2\\
            \cline{2-15}
        \end{tabular}}
        \caption{\label{iclresults} Count number of $(\Nnq,\Nnl)$ models selected by the ICL criterion among 20 data matrices for different difficulties, as measured by the conditional Bayes risk, and different matrix sizes.
        All matrices are generated with the same number of row and column classes: $\Nnq=\Nnl=3$.
        }
    \end{figure}

    \subsection{Analysing the ability of the model selection criterion to select the adequate missingness model}
    \label{sec:icl_exp_missing_model}
    
    %\begin{figure}[h]
    %    \scalebox{0.7}{\input{img/icl_increasing_nmar.tex}}
    %    \caption{ICL with respect to the increasing effect of NMAR missingness (higher is better). \textcolor{orange}{$\star$} is the median error when trained with the MAR model algorithm. \textcolor{green!50!black}{$\triangle$} is the median error when trained with the NMAR model algorithm.}
	%	\label{fig:icl_mar_nmar_2}
    %\end{figure}
    We use the models fitted in Section~\ref{sec:expnmar} to analyze the ability of the ICL criterion to select the right missingness model (\NMARacro or MAR). 
    The difference in ICL between the MAR and \NMARacro models is computed for each data matrix, assuming that the right numbers of classes $(\Nnq,\Nnl)$ are known. 
    
\begin{figure}[tb]
    \centering
        \scalebox{0.7}{\input{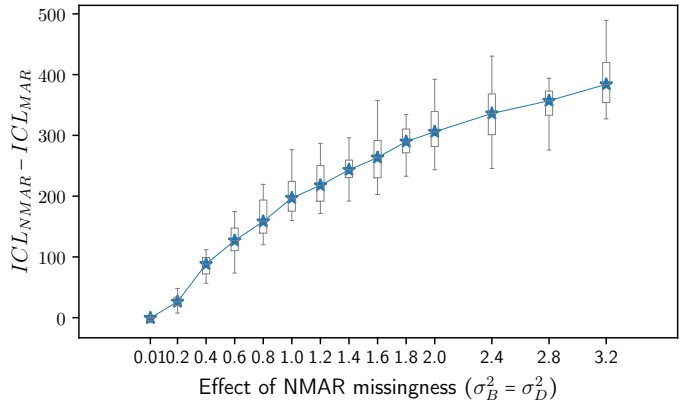}}
        \caption{Difference in ICL between the MAR and \NMARacro models with respect to an increase of the \NMARacro effect, \textcolor[rgb]{0.121569,0.466667,0.705882}{$\filledstar$} is the median. The \NMARacro model is selected when the difference in ICL is positive.}
		\label{fig:icl_mar_nmar}
    \end{figure}

    The results, presented in Figure~\ref{fig:icl_mar_nmar}, show that ICL rightfully opts for the \NMARacro model almost everywhere, demonstrating the ability of this criterion to select the adequate missingness model.
    The MAR model is only chosen for some experiments with the lowest  \NMARacro effect ($\NsigmaB=\NsigmaQ=0.01$), where the prediction performances are almost identical (see Figure \ref{fig:error_classif_nmar}), with a  median difference in ICL of -0.51 (the \MARacro model is chosen 13 times over the 20 repetitions).

\section{Experiments on real data}
    \label{sec:realdata}
    We consider voting records\footnote{Votes from the French National Assembly are available from \url{http://data.assemblee-nationale.fr/travaux-parlementaires/votes}.} from the lower house of the French Parliament (\textit{Assemblée Nationale}). This dataset gathers the results of the 1256 ballots of year 2018 of the 577 French members of parliament (MPs) for the procedural motions and amendments for the 15th legislature (June 2017). 
    For each text, the vote of each MP is recorded as a 4-level categorical response: ``yes'', `no'', ``abstained'' or ``absent''.
    Using our model, we bring out some relevant groups of texts and MPs, as well as some structure in the behavior of nonvoters. 
    
    We gather the data in a matrix where each row represents an MP and each column represents a text. 
    To use our model, we reduced the 4 response levels to 3 (``yes'', `no'', ``missing'') assuming that merging the ``abstained'' and ``absent'' categories would not affect much the underlying missingness process (``abstained'' votes represent about 4\% of the expressed votes, ``missing'' responses represent 85\% of all votes).

\begin{figure}%[hbt!]
        \centering
        \resizebox{0.45\textwidth}{!}{\begin{tikzpicture}.
    \definecolor{color-ni}{RGB}{114, 119, 125}
    \definecolor{color-lr}{RGB}{0, 102, 204}
    \definecolor{color-udi}{RGB}{0, 255, 254}
    \definecolor{color-modem}{RGB}{255, 153, 28}
    \definecolor{color-larem}{RGB}{216, 201, 30}
    \definecolor{color-lt}{RGB}{225, 165, 225}
    \definecolor{color-soc}{RGB}{255, 128, 128}
    \definecolor{color-gdr}{RGB}{203, 36, 67}
    \definecolor{color-fi}{RGB}{221, 0, 19}
    \draw   (0:-5) arc (180:0:5) -- cycle;
    \foreach \a / \b / \color in {0/4.06/color-ni,   4.06/36.6/color-lr, 36.6/45.3/color-udi, 45.3/59.7/color-modem, 59.7/154.7/color-larem, 154.7/160.62/color-lt, 160.62/169.7/color-soc, 169.7/174.7/color-gdr, 174.7/180/color-fi} {
        \draw (0,0) -- (\a:5) arc(\a:\b:5) -- cycle;
        \fill[color=\color]   (0,0) -- (\a:5) arc(\a:\b:5) -- cycle;
        % \fill[pattern=crosshatch dots,pattern color=\color]   (0,0) -- (\a:5) arc(\a:\b:5) -- cycle;
    }
    \node at (5.7,0.2) {\Huge NI};
    \node at (5.7,1.5) {\Huge LR};
    \node at (6.2,3.5) {\Huge UDI-AGIR};
    \node at (5,4.5) {\Huge MODEM};
    \node at (-3.3,5.2) {\Huge LaREM};
    \node at (-5.2,2.6) {\Huge LT};
    \node at (-5.9,1.7) {\Huge SOC};
    \node at (-6.2,0.8) {\Huge GDR};
    \node at (-5.7,-0.1) {\Huge FI};
%    \node[draw, fill=white] at (-5.7,-0.2) {\Huge FI};
\end{tikzpicture}
% # 'NI', Non-inscrits : 13-->4.06
% # 'LR', Les Républicains: 104 -->32.5
% # 'UDI-AGIR', Les Constructifs: 28--> 8.75
% # 'MODEM', Mouvement démocrate : 46--> 14.375
% # 'LaREM, La République En Marche': 304-->95
% # 'LT': 19--> 5.9375
% # 'SOC': 29--> 9.06
% # 'GDR': 16-->5
% # 'FI': 17--> 5.3125
}
        \begin{minipage}[b]{.5\textwidth}
			\baselineskip=0.5\baselineskip
            {\tiny\begin{center}
                Political groups from left-wing to right-wing
            \end{center}
            FI (17): France Insoumise \\
            GDR (16): Groupe de la Gauche démocrate et républicaine \\
            SOC (29): Socialistes \\
            LT (19): Libertés et territoires \\
            LaREM (304): La République En Marche \\
            MODEM (46): Mouvement démocrate\\
            UDI-AGIR (28): Les Constructifs\\
            LR (104): Les Républicains \\
            NI (13): Non inscrits (mixed left and right wings)
            }
        \end{minipage}
        % # 'NI', Non-inscrits: 13-->4.06
        % # 'LR', Les Républicains: 104 -->32.5
        % # 'UDI-AGIR', Les Constructifs: 28--> 8.75
        % # 'MODEM', Mouvement démocrate: 46--> 14.375
        % # 'LaREM, La République En Marche': 304-->95
        % # 'LT': 19--> 5.9375
        % # 'SOC': 29--> 9.06
        % # 'GDR': 16-->5
        % # 'FI': 17--> 5.3125
    \caption{Hemicycle of the political groups of the French National Assembly}
    \label{fig:hemicycle}
    \end{figure}
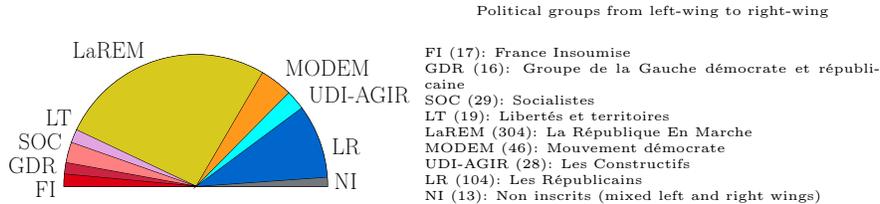

    At the lower house of French Parliament, MPs may group together according to their political affinities.
    Groups with less that 15 members or MPs who choose to be independent are gathered under the ``Non inscrits'' (NI) label, giving a heterogeneous range of political hues inside it. 
    The names of the groups and their frequency are detailed in Figure~\ref{fig:hemicycle}. %The group LaREM is refereed as the majority government as the group possesses the absolute majority of seats in legislature. 
    
\begin{figure}[hbt!]
    \centering
        \includegraphics[width=1.\textwidth]{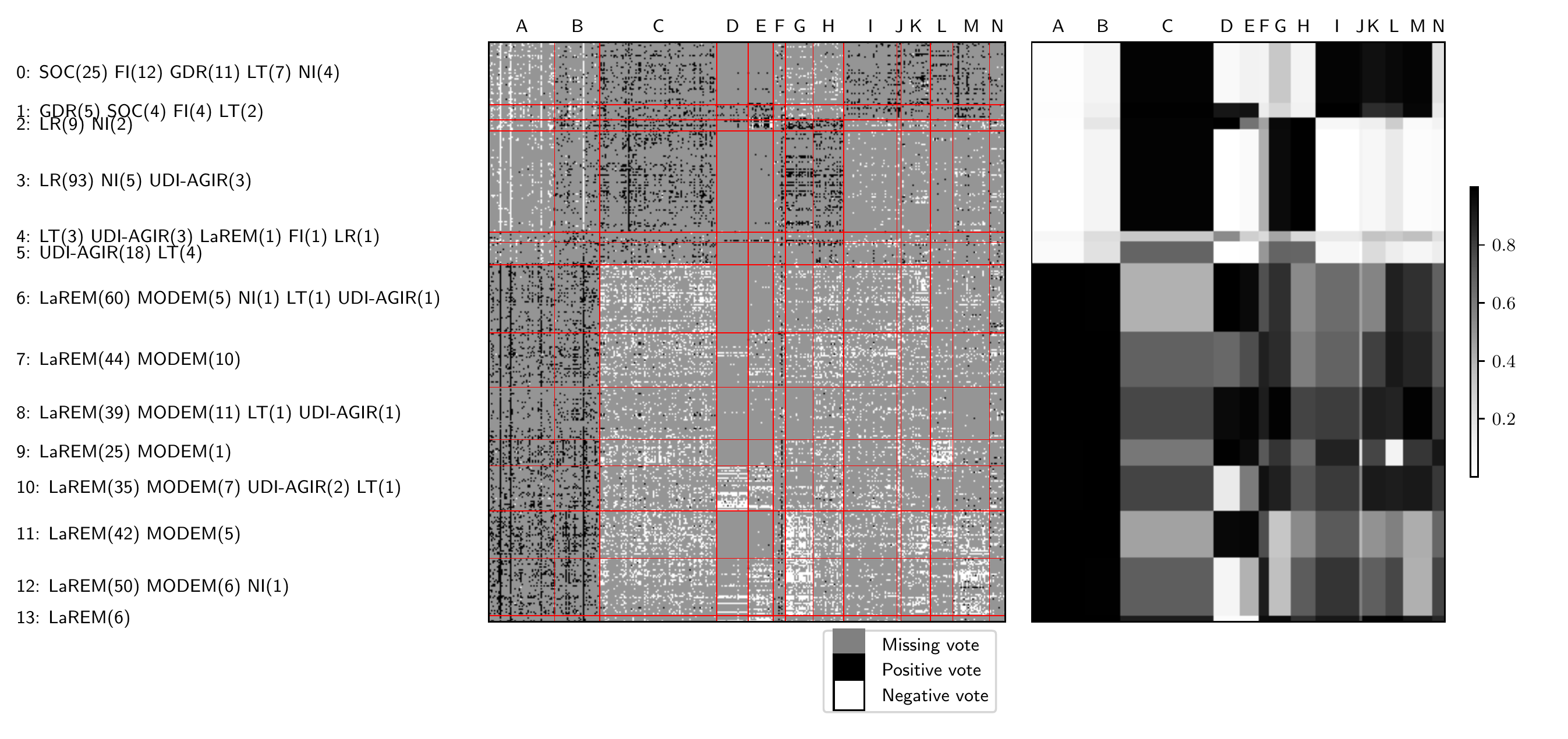}
        \caption{Left: matrix of votes reordered according to the row and column classes, for the \NMARacro LBM model selected by ICL, with 14 MP classes and 14 text classes. The red lines delineate class boundaries. The counts of MPs belonging to their political groups in each MP class is given on the left. 
        Right: summary of the inferred opinions (expressed or not) for all classes of texts and MPs, as given by the estimated probability to support a text in each block of the reordered matrix.
        }
        \label{fig:reordered_asnt}
    \end{figure}

    The ICL criterion, used to  select both the numbers of classes and the type of missingness, favors a \NMARacro missingness with $\Nnq=14$ MP classes and $\Nnl=14$ text classes against a MAR model with 19 MP classes 23 text classes.
    %-259658.4747794035, 19, 23
    %-252498.6874637016, 14, 14
    %The model selected by the ICL is made of $\Nnq=14$ MP classes and $\Nnl=14$ text classes.
    The reordered data matrix derived from this block clustering is displayed in Figure~\ref{fig:reordered_asnt}. 
    Fewer classes lead to over-aggregated components hiding the subtleties of the network, but since they still correspond to well-identified groups and are more friendly to visual analysis, we provide them as additional material in Appendix~\ref{annex:asnt}.
    
    In Figure~\ref{fig:reordered_asnt}, classes of MPs are coherent to their political orientation: class 0 and 1 are mainly made up of left-wing MPs from the groups SOC, FI, GDR, LT, classes 2 and 3 are mainly made up of right-wing MPs from LR and the classes from 6 to 13 are mainly made up of centrist MPs from LaREM and MODEM who are known to be political allies. 
    Classes of texts can be analyzed with the available metadata. 
    A bipartite opposition system appears from classes A and C. 
    Texts from class A are the original articles of law proposed by the government and are unsurprisingly voted positively by the MPs classes from 6 to 13 as they are from the same political mould as the French government. 
    Texts from class C are mainly amendments proposed by minority and are voted positively by both the left wing (class 0 and 1) and the right wing (classes 2 and 3) and negatively by the MPs supporting the government (classes 6 to 13). 
    The left and right wings are yet divided by usual issues such as immigration regulation amendments gathered in classes G and M or general economic matters gathered in classes H and I.
    %A : articles originaux déposés par le gouvernement
    %G : amendements concernant les projets de loi pour une immigration maitrisée ou sur l'accueil des gens du voyage (91\% des textes). Déposés par les Républicains à 93\%.
    % M: "immigration", "gens du voyage", "asile" : déposé par la France insoumise à 70 \%.
    % : I finance, économie générale déposé par la gauche
    %H : idem mais déposé par la droite
    %\begin{figure}[hbt!]
    %    \includegraphics[width=1.\textwidth]{img/plots_anst_13_11_all.pdf}
    %    \caption{Left: reordered voting matrix. Right: probability to vote positively. Red lines are drawn for an improved visualization of classes. The ICL criterion selected 13 MP classes and 11 text classes.}
    %    \label{fig:reordered_asnt}
    %\end{figure}

    %\begin{table}[h!]
    %    \centering
    %    \begin{center}
    %        \begin{tabular}{|c | c c c c|} 
    %        \hline
    %        $(\Nnq, \Nnl)$ & $\NsigmaA$ & $\NsigmaB$& $\NsigmaP$  $\NsigmaP$\\ [0.5ex] 
    %        \hline\hline
    %        (3,5) & 1.28 & 0.05 & 0.88 & 0.23 \\
    %        \hline
    %        (13,11) & 1.25 & 0.51 & 1.02 & 1.19 \\
    %        \hline
    %        (14,14) & 2.74 & 4.83 & 2.34 & 1.11 \\
    %        \hline
    %        \end{tabular}
    %    \end{center}
    %    \caption{Parameters learned by our missingness model. $\NsigmaA$ and $\NsigmaB$
    %are related to MP propensity while $\NsigmaP$ and $\NsigmaQ$ are related to text
    %propensity}
    %\label{table:sigmas}
    %\end{table}
    
\begin{figure}%[hbt!]
    \centering
        \begin{minipage}{0.5\textwidth}
        \includegraphics[width=1.\textwidth]{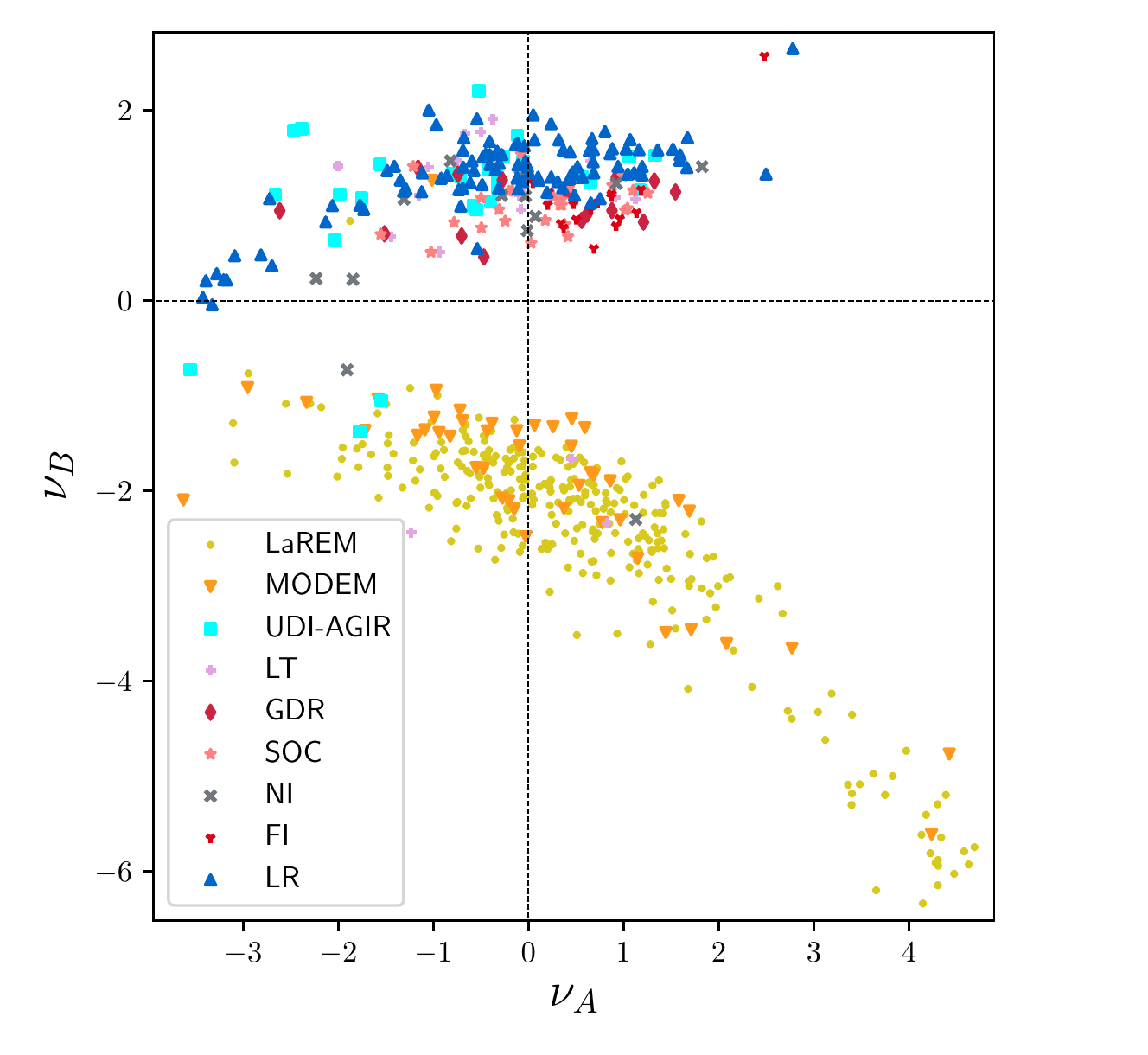}
        \end{minipage}
        \begin{minipage}{0.49\textwidth}
        \includegraphics[width=1\textwidth]{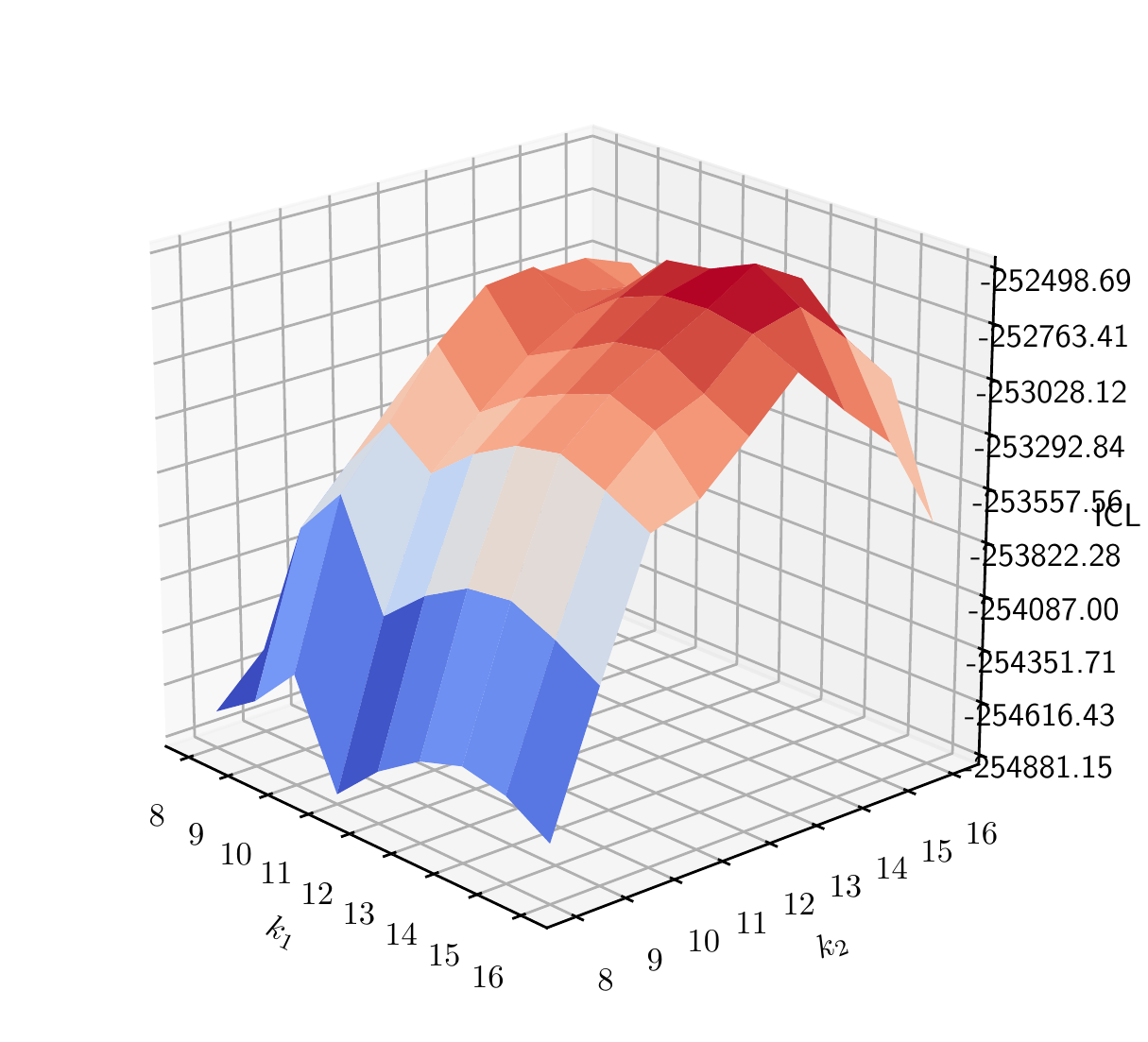}
        \end{minipage}
    \captionof{figure}{Left: maximum \textit{a posteriori} estimates of the MPs propensities ($\NnuA_\Ni$, $\NnuB_\Ni$), with their political group memberships. $\NnuA_\Ni$ drives the MAR effect and $\NnuB_\Ni$ drives the \NMARacro one. Right: ICL curve. Maximum is reached for $\Nnq$=14 and $\Nnl$=14}
    \label{fig:nu_a_nu_b_icl}
    \end{figure}

    %Table~\ref{table:sigmas} displays the trained parameters of our missingness model with a small number of classes (3,5) and the number of classes selected by the ICL (13,11).
    %We remind that the parameters $\NsigmaA$ and $\NsigmaB$ of our model are related to the dispersion of MP propensity and that $\NsigmaB$ drives the MNAR effect. 
    %Similarly, $\NsigmaP$ and $\NsigmaQ$ are related to the dispersion of text propensity with $\NsigmaQ$ driving the item MNAR effect.
    %When using a higher number of classes, the model is more flexible and it is able to detects a stronger MNAR effect imputable on both the MPs and texts. 
    In our model, the latent variables $\NAb$ and $\NBb$ characterize the propensity of MPs to cast a vote.  
    Figure~\ref{fig:nu_a_nu_b_icl} displays the scatter plot of $\NnuA_\Ni$ and $\NnuB_\Ni$, the maximum \textit{a posteriori} estimates of $\NA_\Ni$ and $\NB_\Ni$ for all MPs under the variational distribution. The abscissa represents the propensity to vote\footnote{%
    More rigorously, the abscissa represents the {\em global deviation from the average} propensity to vote.
    }, with higher values of $\NnuAb$ corresponding to a higher propensity to vote, and the ordinate $\NnuBb$ represents the additional effect of casting a vote when supporting the text.
    The membership of MPs to their political group is indicated by the plotting symbol.
    
    We see two obvious clusters separated by the vertical axis $\NnuBb$: the bottom cluster is essentially formed by MPs from the LaREM and MODEM political groups, which support the government, whereas the top cluster is formed by the opposition political groups. 
    %
    %This behavior is not surprising as these MPs are outnumbered by the lower group supporting the government; they consequently have to be more present at the parliament relatively to their opponents if they want to pass a motion or an amendment they deposed.
    The $\NnuBb$ estimates for the opposition cluster are positive, meaning that these MPs come to parliament to vote positively.
    This behavior is not surprising because the MPs of the opposition parties are outnumbered by the MPs supporting the government, so they must be diligent if they want their tabled  motion or amendment passed.
    The dependency between the political groups and the \NMARacro effect encoded in the estimates $\NnuBb$, which is confirmed by an ANOVA test (with a p-value smaller than numerical error), supports that the missingness patterns captured by our model are relevant for the problem at hand. 
    A similar analysis is developed on texts in Appendix \ref{annex:asnt}.
 
    %\begin{figure}[hbt!]
    %    \includegraphics[width=.9\textwidth]{img/nu_b.pdf}
    %    \caption{Histograms of $\NnuB_\Ni$, the maximum a posteriori estimates of the additional MPs propensities to vote positively, grouped according to the political group memberships.}
    %    \label{fig:nu_b}
    %\end{figure}

\section{Conclusion}

In many estimation problems, the absence of data conveys some information on the underlying phenomenon that should be exploited for its modeling.
We propose a co-clustering model that accounts for this absence of data; it aims at retrieving groups of rows and columns based on the complete data matrix instead of considering only the partitioning of the observed data matrix.
This model consists of two building blocks: a co-clustering model (\LBM) of the full data matrix, and a missingness model that manages the censoring that produces the observed data matrix.
This missingness model preserves the symmetry of the co-clustering model by allowing two \NMARacro effects, one on the rows and the other on the columns.
The overall model of the observed data matrix results from the combination of the model of the complete data matrix with the missingness model. 

We used variational techniques and the Delta method to obtain a tractable approximation of the lower bound of the observed log-likelihood.
We proposed a model selection criterion to select both the number of classes and the type of missingness (\MARacro versus \NMARacro). 

Our experiments on synthetic datasets show that ignoring an informative missingness can lead to catastrophic co-clustering estimates, supporting the value of using expressive missingness models on such type of data.
We also illustrate the use of our model on a real-world case where the missingness model provides an interesting basis for analyzing and interpreting the motivations of nonvoters.

Our model should also be useful in other fields such as in ecology, where the probability of observing interaction between species derives from some factors that also explain the true interactions \citep{vazquez2009uniting}, or in collaborative filtering, where the probability of observing a rating depends on the actual rating that would be given by the user \citep{Marlin07}. 
%For the latter application, 
In the latter application, the data sizes generally encountered in recommendation would require computational improvements in inference. 
Another useful future work is to extend our model to non-binary data.
%In the experiment Section~\ref{sec:realdata}, we assumed that the missingness is time independent which is strongly disputable. Thus, in future works, we believe it is necessary to extend the missing data model with time component.

\bibliographystyle{plainnat}
\bibliography{article}

\clearpage
\appendix

\section{Computing the criterion $\mathcal{J}\p{\NRX, \Ntheta}$}\label{annexcriteria}
The criterion to be optimized is :
\begin{equation}
    %\scriptsize
    \mathcal{J}\p{\NRX, \Ntheta} = \mathcal{H}(\NRX) + \expectation_{\NRX}[\llikli(\NXob, \NYoneb, \NYtwob, \NAb, \NBb, \NPb, \NQb; \Ntheta)]
    \enspace,
\end{equation}
where $\Ntheta$ is the list of all model parameters: $\Ntheta=(\Nalphaoneb, \Nalphatwob, \Npib,  \Nmu,\NsigmaA,\NsigmaB,\NsigmaP,\NsigmaQ)$.

We restrict the form of the variational distribution $\NRX$ to get a fully factorized form:
{%\scriptsize
\begin{multline*}
        \NRX=
        \prod_{\Ni=1}^{\Nnone}{\multinomial{1}{\Ntauoneb_\Ni}} \times
        \prod_{\Nj=1}^{\Nntwo}{\multinomial{1}{\Ntautwob_\Nj}}
        \times
        \prod_{\Ni=1}^{\Nnone}{\norm{\NnuA_\Ni}{\NrhoA_\Ni}} \times \\
        \prod_{\Ni=1}^{\Nnone}{\norm{\NnuB_\Ni}{\NrhoB_\Ni}} 
        \times
        \prod_{\Nj=1}^{\Nntwo}{\norm{\NnuP_\Nj}{\NrhoP_\Nj}} \times
        \prod_{\Nj=1}^{\Nntwo}{\norm{\NnuQ_\Nj}{\NrhoQ_\Nj}}
	\enspace,
\end{multline*}}%
\begin{sloppypar}
where $\gamma$ denotes the list of parameters of the variational distribution: %\footnote
\mbox{$\gamma = (\Ntauoneb, \Ntautwob,\NnuAb,\NrhoAb, \NnuBb, \NrhoBb, \NnuPb, \NrhoPb, \NnuQb, \NrhoQb )$}.
The entropy is additive across independant variables, so we get:
\end{sloppypar}
{%\scriptsize
\begin{multline*}
    \mathcal{H}\p\NRX =
        - \sum_{\Ni\Nq}{ \Ntauone_{\Ni\Nq} \log\Ntauone_{\Ni\Nq}}
        - \sum_{\Nj\Nl}{ \Ntautwo_{\Nj\Nl} \log\Ntautwo_{\Nj\Nl}}
        + \p{\Nnone+\Nntwo}\p{\log\p{2\pi}+1} \\
        + \frac{1}{2} \sum_{\Ni}{\p{\log\NrhoA_\Ni+\log\NrhoB_\Ni}}
        + \frac{1}{2} \sum_{\Nj}{\p{\log\NrhoP_\Nj+\log\NrhoQ_\Nj}} 
\enspace.
    %\label{appendix:equ:entropy}
\end{multline*}}

The independence of the latent variables allows to rewrite the expectation of the complete log-likelihood as:
\begin{multline*}
    %\scriptsize
    \expectation_{\NRX}\brackets{\llikli\p{\NXob, \NYoneb, \NYtwob, \NAb, \NBb, \NPb, \NQb}}  =
    \expectation_{\NRX}\brackets{\llikli\p{ \sachant{\NXob}{\NYoneb, \NYtwob, \NAb,  \NBb, \NPb, \NQb}}} \\
    + \expectation_{\NRX}\brackets{\llikli\p{\NYoneb}}
    + \expectation_{\NRX}\brackets{\llikli\p{\NYtwob}} 
    + \expectation_{\NRX}\brackets{\llikli\p{\NAb}}
    + \expectation_{\NRX}\brackets{\llikli\p{\NBb}}  \\
    + \expectation_{\NRX}\brackets{\llikli\p{\NPb}}
    + \expectation_{\NRX}\brackets{\llikli\p{\NQb}}
    \enspace,   
\end{multline*}
with the following terms:
{%\scriptsize 
\begin{align*}
    \expectation_{\NRX} \brackets{\llikli\p{\NYoneb}} & =  \sum_{\Ni\Nq}{\expectation_{\NRX}\NYone_{\Ni\Nq} \log\Nalphaone_\Nq  } = \sum_{\Ni\Nq} { \Ntauone_{\Ni\Nq} \log\Nalphaone_\Nq } \\
    \expectation_{\NRX}\brackets{ \llikli\p{\NYtwob}} & = \sum_{\Nj\Nl}{\expectation_{\NRX}\NYtwo_{\Nj\Nl} \log\Nalphatwo_\Nl  }  = \sum_{\Nj\Nl} { \Ntautwo_{\Nj\Nl} \log\Nalphatwo_\Nl } \\
%\end{align*} }%
%{\scriptsize
%\begin{align*}
    \expectation_{\NRX} \brackets{\llikli\p{\NAb} }&= - \frac{\Nnone}{2} \log2\pi - \frac{\Nnone}{2} \log \NsigmaA - \frac{1}{2\NsigmaA} \sum_{\Ni}{\expectation_{\NRX}\NA_i^2} \\
    &= - \frac{\Nnone}{2} \log2\pi - \frac{\Nnone}{2} \log \NsigmaA - \frac{1}{2\NsigmaA} \sum_{\Ni}\p{\p{\NnuA_\Ni}^2 + \NrhoA_\Ni} \\
    \expectation_{\NRX} \brackets{\llikli\p{\NBb}} &= - \frac{\Nnone}{2} \log2\pi - \frac{\Nnone}{2} \log \NsigmaB - \frac{1}{2\NsigmaB} \sum_{\Ni}\p{\p{\NnuB_\Ni}^2 + \NrhoB_\Ni} \\
    \expectation_{\NRX}\brackets{ \llikli\p{\NPb}} &= - \frac{\Nntwo}{2} \log2\pi - \frac{\Nntwo}{2} \log \NsigmaP - \frac{1}{2\NsigmaP} \sum_{\Nj}\p{\p{\NnuP_\Nj}^2 + \NrhoP_\Nj} \\
    \expectation_{\NRX} \brackets{\llikli\p{\NQb}} &= - \frac{\Nntwo}{2} \log2\pi - \frac{\Nntwo}{2} \log \NsigmaQ - \frac{1}{2\NsigmaQ} \sum_{\Nj}\p{\p{\NnuQ_\Nj}^2 + \NrhoQ_\Nj}
\end{align*}}%
and
\iffalse
\begin{multline}\label{appendix:equ:r}
    % R_ij
    \expectation_{\NRX}\brackets{\llikli\p{\sachant{\NXob}{\NYoneb, \NYtwob, \NAb, \NBb, \NPb, \NQb}}} = \\
    %positives
    \sum_{\Nq\Nl,\Ni\Nj:\NXo_{\Ni\Nj}=1} \hspace{-10pt} {\Ntauone_{\Ni\Nq}\Ntautwo_{\Nj\Nl} \expectation_{\NRX}\brackets{\log\p{\pi_{\Nq\Nl} \logistic{\Noddp}}} } \\
    % negatives
    +  \hspace{-10pt} \sum_{\Nq\Nl,\Ni\Nj:\NXo_{\Ni\Nj}=0} \hspace{-10pt} {\Ntauone_{\Ni\Nq}\Ntautwo_{\Nj\Nl} \expectation_{\NRX}\brackets{\log\p{ (1-\pi_{\Nq\Nl}) \logistic{\Noddm} }} } \\
    % NA
    +  \hspace{-10pt} \sum_{\Nq\Nl,\Ni\Nj:\NXo_{\Ni\Nj}=\NNA} \hspace{-20pt} {\Ntauone_{\Ni\Nq}\Ntautwo_{\Nj\Nl} \expectation_{\NRX}\brackets{\log\p{1-\pi_{\Nq\Nl} \logistic{p_1} -(1-\pi_{\Nq\Nl}) \logistic{p_2} }} }
	\enspace,
\end{multline}
\fi
\begin{multline}\label{appendix:equ:r}
    % R_ij
    \expectation_{\NRX}\brackets{\llikli\p{\sachant{\NXob}{\NYoneb, \NYtwob, \NAb, \NBb, \NPb, \NQb}}} = 
    %positives
    \sum_{\Nq\Nl,\Ni\Nj:\NXo_{\Ni\Nj}=1} \hspace{-10pt} {\Ntauone_{\Ni\Nq}\Ntautwo_{\Nj\Nl} \expectation_{\NRX}\brackets{\log p_1} } \\
    % negatives
    +  \hspace{-10pt} \sum_{\Nq\Nl,\Ni\Nj:\NXo_{\Ni\Nj}=0} \hspace{-10pt} {\Ntauone_{\Ni\Nq}\Ntautwo_{\Nj\Nl} \expectation_{\NRX}\brackets{\log p_0} } 
    % NA
    +  \hspace{-10pt} \sum_{\Nq\Nl,\Ni\Nj:\NXo_{\Ni\Nj}=\NNA} \hspace{-20pt} {\Ntauone_{\Ni\Nq}\Ntautwo_{\Nj\Nl} \expectation_{\NRX}\brackets{\log\p{1-p_0 -p_1}} }
	\enspace,
\end{multline}
with $p_0$ and $p_1$ defined in Equations \eqref{eq:p2}--\eqref{eq:p1}.

Equation~\eqref{appendix:equ:r} involves the computation of the expectations of the following nonlinear functions:
{%\scriptsize
\begin{equation*}
    \begin{array}{l}
        f_1(x,y) = \log\p{\pi_{\Nq\Nl} \logistic\p{\Nmu + x + y} }\\
        f_0(x,y) = \log\p{ (1-\pi_{\Nq\Nl}) \logistic\p{\Nmu + x - y} }\\
        f_\NNA(x,y) = \log\p{1-\pi_{\Nq\Nl} \logistic\p{\Nmu + x + y} -(1-\pi_{\Nq\Nl}) \logistic\p{\Nmu + x - y} }
	    \enspace.
    \end{array}
\end{equation*}}
The approximation of these expectations given by the second-order Delta method with independent random variables  $X$ and $Y$ reads:
{%\scriptsize 
\begin{multline*}
    \expectation\brackets{f\p{X,\; Y}}  \approx f\p{\expectation X,\; \expectation Y}
    + \frac{1}{2} \; var(X) \; \frac{\partial^2  f\p{\expectation\brackets{X},\;\expectation \brackets{Y}}}{\partial \p{X}^2} \\
    + \frac{1}{2} \; var(Y) \; \frac{\partial^2  f\p{\expectation\brackets{X},\; \expectation\brackets{ Y}}}{\partial \p{Y}^2}
	\enspace,
\end{multline*}}
which yields in our case:
{%\scriptsize
\begin{multline*}
    \expectation_{\NRX}\brackets{f\p{\NA_\Ni+\NP_\Nj,\; \NB_\Ni+\NQ_\Nj}}  \approx f\p{\NnuA_\Ni+\NnuP_\Nj,\; \NnuB_\Ni+\NnuQ_\Nj}\\
    + \frac{1}{2} \p{\NrhoA_\Ni +\NrhoP_\Nj} \; \frac{\partial^2  f\p{\NnuA_\Ni+\NnuP_\Nj,\; \NnuB_\Ni+\NnuQ_\Nj}}{\partial \p{\NnuA_\Ni+\NnuP_\Nj}^2} \\
    + \frac{1}{2} \p{\NrhoB_\Ni +\NrhoQ_\Nj} \; \frac{\partial^2  f\p{\NnuA_\Ni+\NnuP_\Nj,\; \NnuB_\Ni+\NnuQ_\Nj}}{\partial \p{\NnuB_\Ni+\NnuQ_\Nj}^2}
	\enspace.
\end{multline*}}
The criterion is now fully computable.

\section{Initialization of the VEM algorithm with spectral clustering.}
\label{annex:init}
    \begin{algorithm}[H]
    \SetAlgoLined
    \textbf{Input}: 
    \begin{itemize}
        \item observed data $\NXob$
        \item $\Nnq$ and $\Nnl$ number of row groups and column groups
    \end{itemize}
    \SetKwFunction{FMain}{SpectralClustering}
    \SetKwProg{Fn}{Function}{:}{}
    \Fn{\FMain{$W$ adjacency matrix, $k$ number of clusters}}{
        \begin{itemize}
            \item Define $\boldsymbol{D}$ the diagonal matrix, element of $\mathbb{R}^{n\times n}$: $D_{\Ni\Ni}=\sum_\Nq W_{\Ni\Nq}$
            \item Define $\boldsymbol{L}=\boldsymbol{D}^{-1/2} \boldsymbol{W} \boldsymbol{D}^{-1/2}$
            \item Find the eigenvectors corresponding to the $k$ eigenvalues of $\boldsymbol{L}$ that are largest in absolute value. Form the matrix $\boldsymbol{U} = [U_1,...,U_k] \in \mathbb{R}^{n\times k}$  concatenating the eigenvectors into columns.
        \end{itemize}
        \textbf{Return} results of \textit{k}-means with $k$ clusters on $\boldsymbol{U}$.
    }
    \Begin{
        \begin{itemize}
        \item Build $\NYoneb$ the $\Nnone\times\Nnq$
            indicator matrix of the row cluster memberships with results of $SpectralClustering(\boldsymbol{X} \boldsymbol{X}^T, \Nnq)$
            
        \item Build $\NYtwob$ the $\Nntwo\times\Nnl$
            indicator matrix of the column cluster memberships with results of $SpectralClustering(\boldsymbol{X}^T \boldsymbol{X}, \Nnl)$.
        \item $\Nalphaoneb$, $\Nalphatwob$ and $\Npib$ estimated from $\NYoneb$ and $\NYtwob$
        \item $\Nmu$ initialized such as $\logistic(\Nmu)$ is the global missingness rate
        \item $\NsigmaA$, $\NsigmaB$, $\NsigmaP$ and $\NsigmaQ$ sampled from $\textit{U}_{]0,1]}$
        \end{itemize}
    }
     \KwResult{
        \begin{itemize}
            \item $\theta=\p{\Nalphaoneb, \Nalphatwob, \Npib, \Nmu, \NsigmaA, \NsigmaB, \NsigmaP, \NsigmaQ}$ the model parameters
            \item $\NYoneb$ and $\NYtwob$ the row and column cluster memberships
        \end{itemize}}
     \caption{Initialization of the VEM algorithm with spectral clustering.}
     \label{algo:init}
    \end{algorithm}

\section{Asymptotic form of the Integrated Completed Likelihood}\label{annexicl}
\subsection{ICL of the \NMARacro model}

The ICL criterion of the \LBMacro extended to the \NMARacro missingness process presented in Section \ref{sect:lbmextended} has the following asymptotic form for $\Nnone$ and $\Nntwo$:
\begin{align}
    ICL(\Nnq, \Nnl) =&
    {\max_{\Nthetab}}\; \log \Prob{\NXob,\NYoneb, \NYtwob, \NAb, \NBb, \NPb, \NQb; \Nthetab}
    - \frac{\Nnq\Nnl}{2} \log\p{\Nnone\Nntwo} \nonumber\\
    & - \frac{\Nnq-1}{2} \log\p{\Nnone} - \frac{\Nnl-1}{2} \log\p{\Nntwo} \label{annex:eq:icl}\\
    & + \Nnone \log\p{2\pi} - \log\p{\Nnone}
    + \Nntwo \log\p{2\pi} - \log\p{\Nntwo} \nonumber\\
    & + o(\log\Nnone) + o(\log\Nntwo) \nonumber
    \enspace.
\end{align}

\begin{proof}
With independent latent variables and independent priors on the parameters, the ICL criterion reads
\begin{align}
    ICL &= \llikli\p{\NXob, \NYoneb, \NYtwob, \NAb, \NBb, \NPb, \NQb} \nonumber\\  
    &= \log \int \Prob{\sachant{\NXob, \NYoneb, \NYtwob, \NAb, \NBb, \NPb, \NQb}{\Nthetab}} \Prob{\Nthetab}\mathrm{d}\Nthetab \nonumber\\
    &= \log\int\Prob{\sachant{\NXob}{\NYoneb,\NYtwob,\NAb,\NBb,\NPb,\NQb,\Npib}}\Prob{\Npib}\Prob{\Nmu}\mathrm{d}\Npib \mathrm{d}\Nmu \label{annex:eq:iclfull}\\
    &\quad+ \log\int\Prob{\sachant{\NYoneb}{\Nalphaoneb}}\Prob{\Nalphaoneb}\mathrm{d}\Nalphaoneb
    + \log\int\Prob{\sachant{\NYtwob}{\Nalphatwob}}\Prob{\Nalphatwob}\mathrm{d}\Nalphatwob \nonumber\\
    &\quad + \log\int\Prob{\sachant{\NAb}{\NsigmaA}}\Prob{\NsigmaA}\mathrm{d}\NsigmaA + \log\int\Prob{\sachant{\NBb}{\NsigmaB}}\Prob{\NsigmaB}\mathrm{d}\NsigmaB\nonumber \\
    &\quad + \log\int\Prob{\sachant{\NPb}{\NsigmaP}}\Prob{\NsigmaP}\mathrm{d}\NsigmaP + \log\int\Prob{\sachant{\NQb}{\NsigmaQ}}\Prob{\NsigmaQ}\mathrm{d}\NsigmaQ\nonumber
    \enspace.
\end{align}
As in the ICL developed by \cite{keribinicl} for the standard \LBMacro, we set non-informative Dirichlet distribution $\mathcal{D}(a,...,a)$ priors on $\Nalphaoneb$ and $\Nalphatwob$:
\begin{align*}
    \log\Prob{\NYoneb}
    &=\log\int\Prob{\sachant{\NYoneb}{\Nalphaoneb}}\Prob{\Nalphaoneb;a}\mathrm{d}\Nalphaoneb \\
    &=\prod_{\Ni}\log\int \prod_{\Nq}{\p{\Nalphaone_\Nq}^{\NYone_{\Ni\Nq}}}   \frac{1}{\mathcal{B}(a)}\prod_{\Ni\Nq}{\p{\Nalphaone_\Nq}^{a-1}}\mathrm{d}\Nalphaoneb \\
    & = \log \mathcal{B}(a+\sum_{\Ni}{\NYoneb_\Ni}) - \log {\mathcal{B}(a)} \\
    &= \sum_\Nq{\log\Gamma(\NYone_{\!:\Nq}+a)}+\log\Gamma(\Nnq a)-\log\Gamma(\Nnone+\Nnq a)-\Nnq\log\Gamma(a)
    \enspace,
\end{align*}
where {$\NYone_{\!:\Nq} = \sum_\Ni\NYone_{\Ni\Nq}$}.
The Stirling approximation {$\log \Gamma(x)= x\log x - x -\frac{1}{2}\log x + o(\log x)$} leads to the following asymptotic development of $\log \Prob{\NYoneb}$:
%\scriptsize
\begin{align*}
        \log \Prob{\NYoneb}
        &=  \sum_\Nq{\log \Gamma(\NYone_{:\Nq}+a) } - \log \Gamma(\Nnone+\Nnq a)  +o(\log \Nnone) \\
        &= \sum_{\Nq}{\NYone_{\!:\Nq}} \log\NYone_{\!:\Nq} - \Nnone - \frac{1}{2}\Nnone \\ &
        \quad - \p{\Nnone \log\Nnone + \Nnq a \log\Nnone - \Nnone -\frac{1}{2}\log\Nnone} +o(\log \Nnone)
    \enspace.
\end{align*}
With the non-informative Jeffrey prior $a=\frac{1}{2}$, this gives:
%\scriptsize
\begin{align}
    \log \Prob{\NYoneb}
    &= \sum_{\Nq}{\NYone_{\!:\Nq} \log(\frac{1}{\Nnone}\NYone_{\!:\Nq})} - \frac{\Nnq-1}{2} \log\Nnone +o(\log \Nnone) \nonumber\\
    &= \underset{\Nalphaoneb}{\max} \log \Prob{\NYoneb; \Nalphaoneb} - \frac{\Nnq-1}{2}\log \Nnone +o(\log \Nnone)
    \label{equ:asymptoticY1}
    \enspace.
\end{align}
Similarly, we get:
%\scriptsize
\begin{align}
    \log \Prob{\NYtwob} &= \sum_\Nl{\log \Gamma(\NYtwo_{:\Nl}+a) } + \log \Gamma(\Nnl a) - \log \Gamma(\Nntwo+\Nnl a) - \Nnl \log \Gamma(a) \nonumber\\
    &= \underset{\Nalphatwob}{\max} \log \Prob{\NYtwob; \Nalphatwob} - \frac{\Nnl-1}{2}\log \Nntwo +o(\log \Nntwo)
    \label{equ:asymptoticY2}
    \enspace,
\end{align}
where {$\NYtwo_{:\Nl} = \sum_\Nj\NYtwo_{\Nj\Nl}$}.

We set non-informative InverseGamma($\beta$, $\beta$) distributions (as $\beta$ tends to zero) as priors on $\NsigmaA$, $\NsigmaB$, $\NsigmaP$ and $\NsigmaQ$:
%\scriptsize
\begin{align*}
    \log\Prob{\NAb}
    &= \int{\Prob{\NAb|\NsigmaA}\Prob{\NsigmaA;\beta}\; \mathrm{d}\NsigmaA}\\
    &= \prod_\Ni\log\int \p{2\NsigmaA}^{-\frac\Nnone 2} \exp\p{-\frac{\sum{\NA_\Ni^2}}{2\NsigmaA}}\; \frac{\beta^\beta}{\Gamma(\beta)} \exp\p{-\frac{\beta}{\NsigmaA}}  \p{\NsigmaA}^{-\beta-1}  \; \mathrm{d}\NsigmaA \\
    &= \log\frac{\beta^\beta}{\Gamma(\beta)} 2^{\p{-\frac{\Nnone}{2}}}\int \NsigmaA{}^{\p{-\frac{\Nnone}{2}-\beta-1}} \exp\p{-\frac{2\beta+\sum{\NA_\Ni^2}}{2}\cdot \frac{1}{\NsigmaA}}  \; \mathrm{d}\NsigmaA \\
    &= \log\frac{\beta^\beta}{\Gamma(\beta)} 2^{\beta} \; \p{2\beta+\sum{\NA_\Ni}}^{\p{-\frac{\Nnone}{2}-\beta}} \; \Gamma(\frac{\Nnone}{2}+\beta)
    \enspace.
\end{align*}
To consider a non-informative InverseGamma($\beta$, $\beta$) distribution, we realize a first order Taylor development as $\beta$ tends to 0:
\begin{equation*}
    %\scriptsize
    \log \Prob{\NAb} \approx \log \Gamma(\frac{\Nnone}{2}) + \log \beta - \frac{\Nnone}{2} \log(\sum{\NA_\Ni^2})
    \enspace.
\end{equation*}
Using the Stirling approximation of $\Gamma(x)$ we get the following asymptotic development of $\log\Prob{\NAb}$:
\begin{align}
    \log\Prob{\NAb}
    &= \frac{\Nnone}{2} \log \Nnone - \frac{\Nnone}{2} - \frac{1}{2} \log \Nnone - \frac{\Nnone}{2} \log \sum{\NA_\Ni^2} +o(\log \Nnone) \nonumber\\
    &= \underset{\NsigmaA}{\max} \log \Prob{\NAb; \NsigmaA} + \frac{\Nnone}{2}\log(2\pi) - \frac{1}{2} \log \Nnone +o(\log \Nnone)
    \enspace.
    \label{equ:asymptoticA}
\end{align}
Similarly, we get:
{%\scriptsize
\begin{align}
    \log \Prob{\NBb} = \underset{\NsigmaB}{\max} \log \Prob{\NBb; \NsigmaB} + \frac{\Nnone}{2}\log(2\pi) - \frac{1}{2} \log \Nnone +o(\log \Nnone) \nonumber\\ 
    \log \Prob{\NPb} = \underset{\NsigmaP}{\max} \log \Prob{\NPb; \NsigmaP} + \frac{\Nntwo}{2}\log(2\pi) - \frac{1}{2} \log \Nntwo +o(\log \Nntwo) \label{equ:asymptoticBPQ} \\
    \log \Prob{\NQb} = \underset{\NsigmaQ}{\max} \log \Prob{\NQb; \NsigmaQ} + \frac{\Nntwo}{2}\log(2\pi) - \frac{1}{2} \log \Nntwo +o(\log \Nntwo) \nonumber
    \enspace.
\end{align}}

Using the standard BIC approximation, we have
\begin{align}
    \log \Prob{\NXob|\NYoneb, \NYtwob}
    &= \log \int \Prob{\NXob|\NYoneb, \NYtwob, \Npib}\Prob{\Npib}\Prob{\Nmu} \mathrm{d}\Npib \mathrm{d}\Nmu\nonumber\\
    &=\; \max_{\Npib} \log \Prob{\NXob|\NYoneb, \NYtwob; \Npib, \Nmu} + \frac{\Nnq\Nnl}{2}\log(\Nnone\Nntwo) \label{equ:bic} \\
    &\quad +  o(\log \Nnone)+o(\log \Nntwo) \nonumber  
    \enspace.
\end{align}
The ICL criterion \ref{annex:eq:icl} is directly derived from equations \ref{annex:eq:iclfull}, \ref{equ:asymptoticY1}, \ref{equ:asymptoticY2}, \ref{equ:asymptoticA}, \ref{equ:asymptoticBPQ} and \ref{equ:bic}.

\end{proof}

\subsection{ICL of the \LBMacro with MAR data}
We consider the following \LBMacro extended with the MAR missingness process:
{%\scriptsize
\begin{align*}
% \forall \Ni,\Nj, \left\{
    &\text{Latent Block Model} \\
    &\quad \NYone_\Ni \iidsim \multinomial1{\Nalphaoneb}, \qquad  \Nalphaoneb \in \mathbf{S}_{\Nnq-1} \\
    &\quad \NYtwo_\Nj \iidsim \multinomial1{\Nalphatwob}, \qquad  \Nalphatwob \in \mathbf{S}_{\Nnl-1} \\
    &\quad \p{\sachant{\NXc_{\Ni\Nj}}{\NYone_{\Ni}=\Nq, \NYtwo_{\Nj} = \Nl}} \indsim \extbernoulli{}{\Npi_{\Nq\Nl}}, \qquad  \Npi_{ql} \in [0,1]\\
    &\text{MAR data model} \\
    &\quad\NA_\Ni \iidsim \norm0{\NsigmaA}, \qquad  \NsigmaA \in \mathds{R}_{+}^{*} \\
    &\quad\NP_\Nj \iidsim \norm0{\NsigmaP}, \qquad  \NsigmaP \in \mathds{R}_{+}^{*} \\
    &\quad\p{\sachant{ M_{\Ni\Nj} }{\NA_\Ni,\NP_\Nj}} \indsim  \bernoulli{\logistic\p{\Nmu+\NA_\Ni +\NP_\Nj}} \\
    &\text{Observations are generated according to:} \\
    &\quad\p{\sachant{ \NXo_{\Ni\Nj} }{ \NXc_{\Ni\Nj}, M_{\Ni\Nj}}} =
    \left\{
      \begin{array}{cll}
        \NXc_{\Ni\Nj} &\text{ if } \quad \NM_{\Ni\Nj} = 1 \\
        \NNA  &\text{ if } \quad \NM_{\Ni\Nj} = 0 \\
      \end{array}
    \right. \nonumber
\end{align*}
}
The ICL of this model has the following asymptotic form for $\Nnone$ and $\Nntwo$:
{%\scriptsize
\begin{align}
    ICL(\Nnq, \Nnl) =\;&
    \max_{\Nthetab}\;\log\Prob{\NXob,\NYoneb,\NYtwob,\NAb, \NPb;\Nthetab}
    - \frac{\Nnq\Nnl}{2}\log\p{\Nnone\Nntwo} \nonumber\\
    & - \frac{\Nnq-1}{2} \log\p{\Nnone} - \frac{\Nnl-1}{2} \log\p{\Nntwo} \label{annex:eq:iclmar}\\
    & + \frac{1}{2}\p{\Nnone \log\p{2\pi} - \log\p{\Nnone}
    + \Nntwo \log\p{2\pi} - \log\p{\Nntwo}} \nonumber\\
    & + o(\log\Nnone) + o(\log\Nntwo) \nonumber
    \enspace.
\end{align}}

\section{Supplemental figures for estimations}
\label{annex:estimation}
   This section provides additional experimental results that show a consistent estimation of the model parameters. We reuse the data matrices generated by the \LBMacro with missing data from Section~\ref{sec:classifinference}.
   An initial data matrix of size $\Nnone=\Nntwo=500$ with a conditional Bayes risk of $5\%$ was generated and progressively reduced, removing rows and columns, to increase the difficulty of the classification task.
    
\begin{figure}[H]
    \centering
        \resizebox{0.5\textwidth}{!}{\input{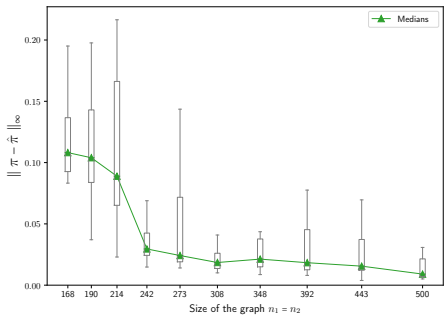}}
        \caption{Maximum error between the true ($\Npib$) and the estimated ($\hat{\Npib}$) probabilities associated to the blocks of the data matrix $\NXcb$ as a function of its size.}  
		\label{fig:infinity_error}
    \end{figure}

    Figure~\ref{fig:infinity_error} displays the maximum absolute error  made on the parameters $\Npib$ of the Bernoulli distributions that model the probability of $\NXcb$ conditionally to the row and column classes. This error decreases as the size of the data matrices grows, which is consistent with our expectations. %We observe a higher variability of the error when the matrix get smaller. 

    Figure \ref{fig:mse_abcd}  displays the mean squared error (MSE) between the generated and estimated values of the latent variables $\NAb$, $\NBb$, $\NPb$, $\NQb$ responsible for the individual variability of missingness. 
    The estimated values are given by the maximum \textit{a posteriori} of their corresponding variational distribution.
    The MSE curves of the variables $\NAb$ and $\NPb$ are comparable as well as the curves of the variables $\NBb$ and $\NQb$. This is expected as the data matrices are generated with symmetric characters in rows and columns.
    \begin{figure}[H]
    \centering
        \resizebox{0.5\textwidth}{!}{\input{img/mse_abcd.pgf}}
        \caption{Mean squared error of the maximum \textit{a posteriori} estimates of the latent variables $\NAb$, $\NBb$, $\NPb$, $\NQb$ governing the propensity of missingness.}  
		\label{fig:mse_abcd}
    \end{figure}
    
    %\begin{figure}[H]
    %    \resizebox{0.8\textwidth}{!}{\input{img/abcd_bissect_500.pgf}}
    %    \caption{Maximum \textit{a posteriori} estimates of the latent variables governing the propensity of missingness versus their true generated values. In this setting: $\Nnone=\Nntwo=500$ and the conditionnal Bayes' risk is 0.05. The identity line is drawned in red for reference.}  
	%	\label{fig:abcd_bissect_500}
    %\end{figure}
    
\begin{figure}[H]
   \centering
        \resizebox{0.9\textwidth}{!}{\input{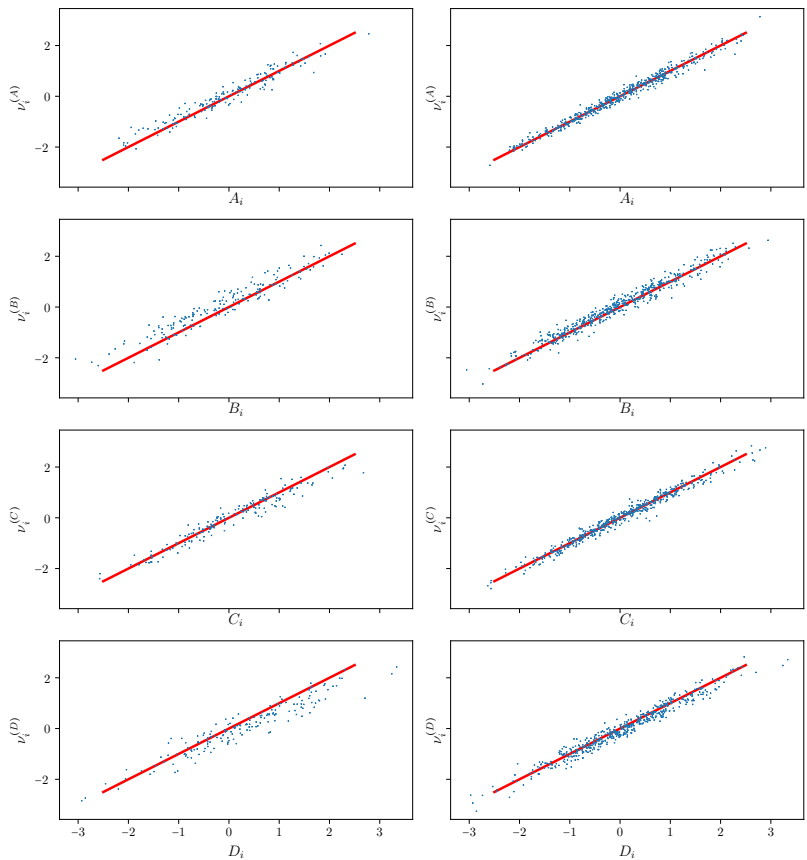}}
        \caption{Maximum \textit{a posteriori} estimates of the latent variables governing the propensity of missingness versus their true generated values.Left: $\Nnone=\Nntwo=168$ and the conditionnal Bayes risk is 0.44;
        right: $\Nnone=\Nntwo=500$ and the conditionnal Bayes risk is 0.05. The identity line is drawn in red for reference.}  
		\label{fig:abcd_bissect_168_500}
    \end{figure}

    Figure~\ref{fig:abcd_bissect_168_500} compares the estimated values of $\NAb$, $\NBb$, $\NPb$ and $\NQb$ to their true generated values for two different sizes of data matrices, all other parameters being equal. 
    A linear trend is exhibited from these scatter plots showing a good aptitude of the proposed inference to recover extreme negative and positives values.

    %\begin{figure}[H]
    %    \resizebox{0.8\textwidth}{!}{\input{img/abcd_bissect_168.pgf}}
    %    \caption{Maximum \textit{a posteriori} estimates of the latent variables governing the propensity of missingness versus their true generated values. In this setting: $\Nnone=\Nntwo=168$ and the conditionnal Bayes' risk is 0.44. The identity line is drawned in red for reference.}  
	%	\label{fig:abcd_bissect_168}
    %\end{figure}

\section{Supplemental figures for the French national assembly votes analysis}
\label{annex:asnt}

\begin{figure}%[H]
\centering
    \includegraphics[height=5cm,trim={0cm 1.75cm 0cm 0cm},clip]{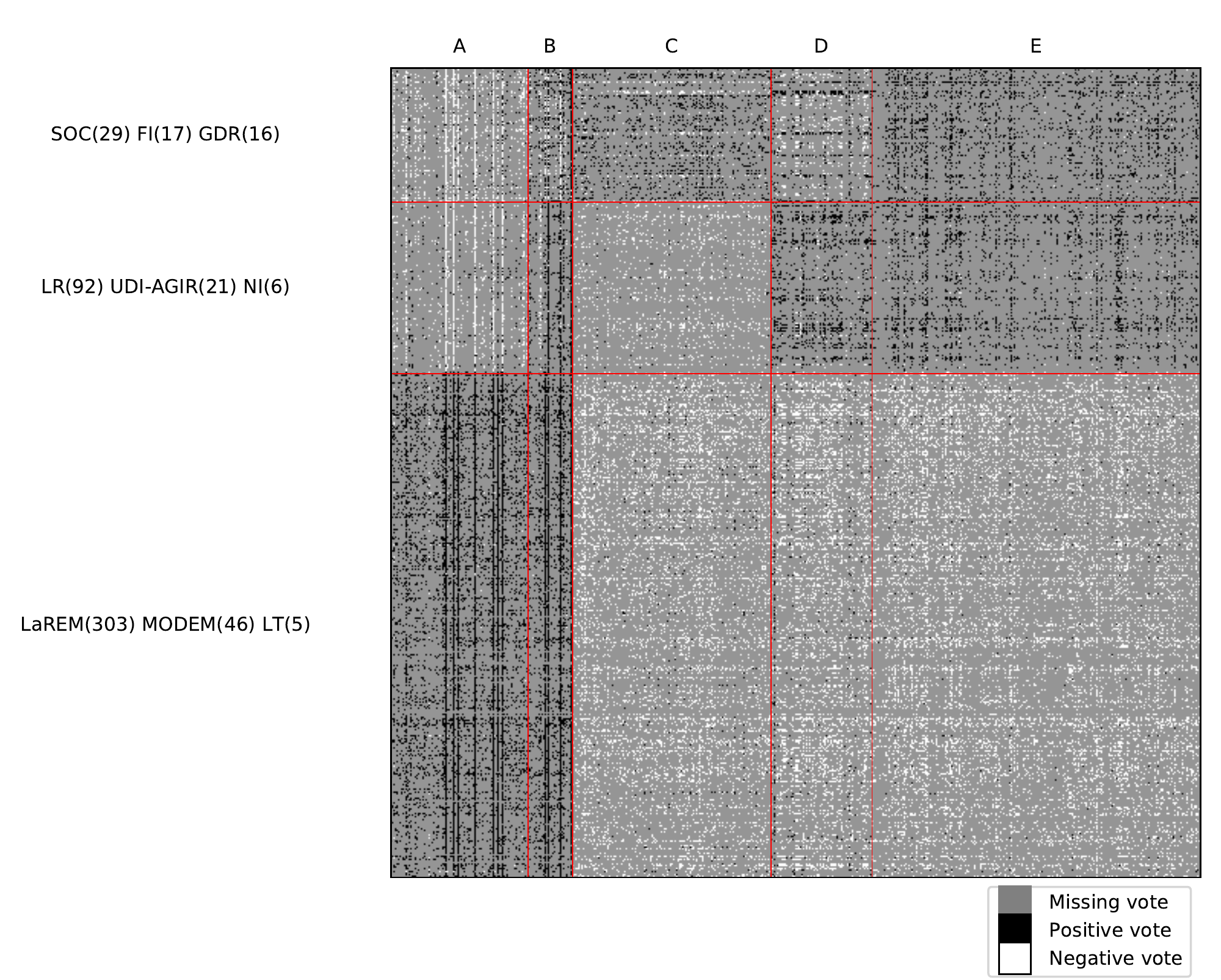}%
    \includegraphics[height=4.85cm,trim={5.5cm 0.6cm 1.5cm 0cm},clip]{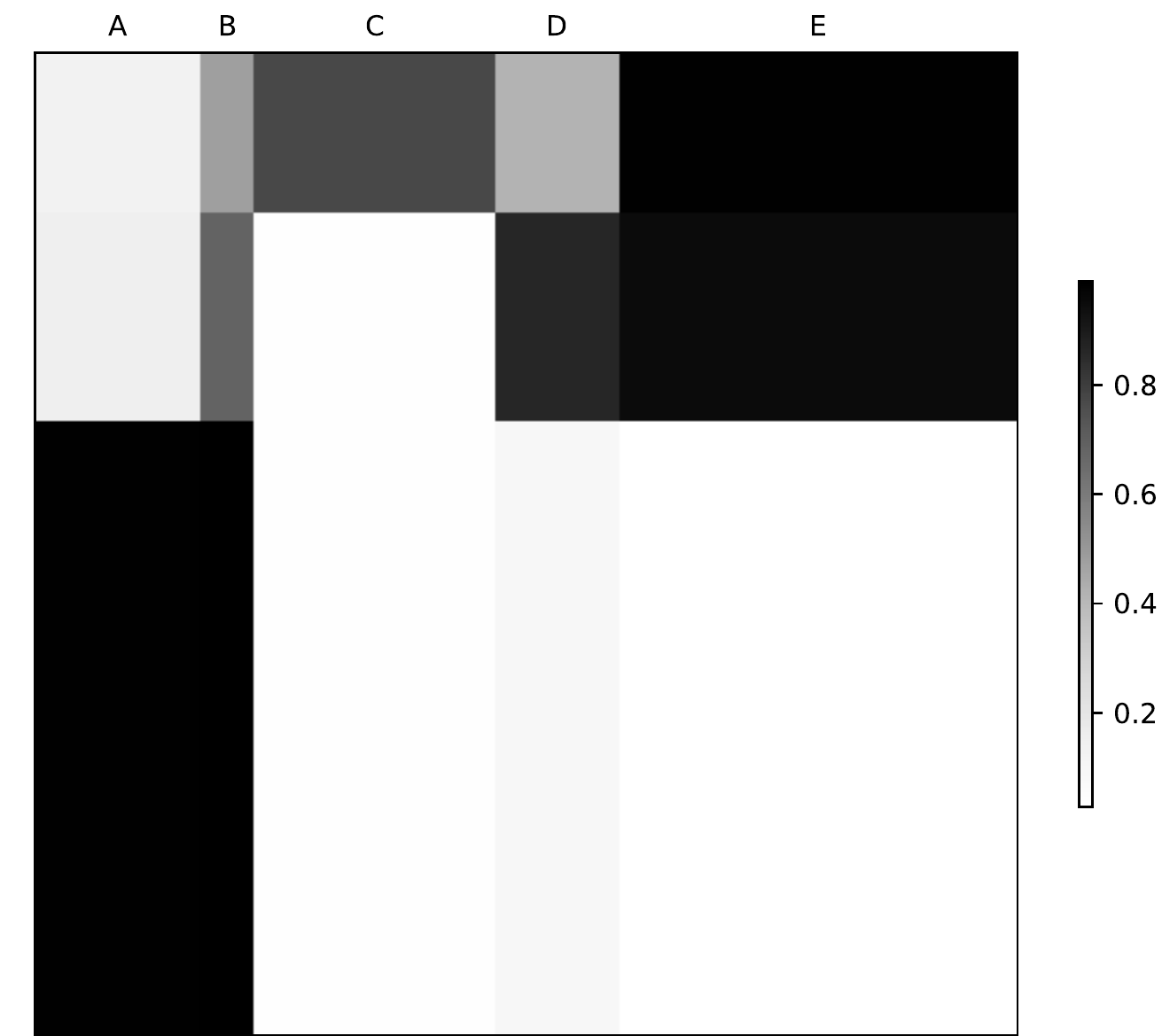}
    \caption{Left: matrix of votes reordered according to the row and column classes, for the MNAR LBM with 3 MP classes and 5 text classes. The red lines delineate class boundaries. The counts of MPs belonging to the three most represented political groups in each MP class is given on the left.
	Right: summary of the inferred opinions (expressed or not) for all classes of texts and MPs, as given by the estimated probability to support a text in each block of the reordered matrix.
}
    \label{fig:reordered_asnt_small}
\end{figure}

Figure~\ref{fig:reordered_asnt_small} displays the reordered matrix of votes derived from a block clustering with a small number of classes. 
Such a simplification may be helpful for identifying global trends.  
With this model, the three MP classes are broadly identified as gathering the right-wing (first class) and left-wing (second class) opposition parties, the last class being formed of the political groups supporting the government.
The opposition systems  appear clearly: on the texts from classes A and E, the votes contrast the membership to the opposition parties versus the governmental alliance, whereas on texts from classes C and D, they separate the left-wing from the right-wing oppositions.
Class B gathers various texts on topics of rather general agreement pertaining to social or health matters.
    
%\begin{figure}[H]
%    \includegraphics[width=0.7\textwidth]{img/3_5_prob.pdf}
%    \caption{Probability to vote positively in each block defined by the Cartesian product of the 3 MP classes and the 5 text classes.}
    %\label{fig:prob_asnt_small}
%\end{figure}

\begin{figure}%[H]
\centering
    \includegraphics[width=0.7\textwidth]{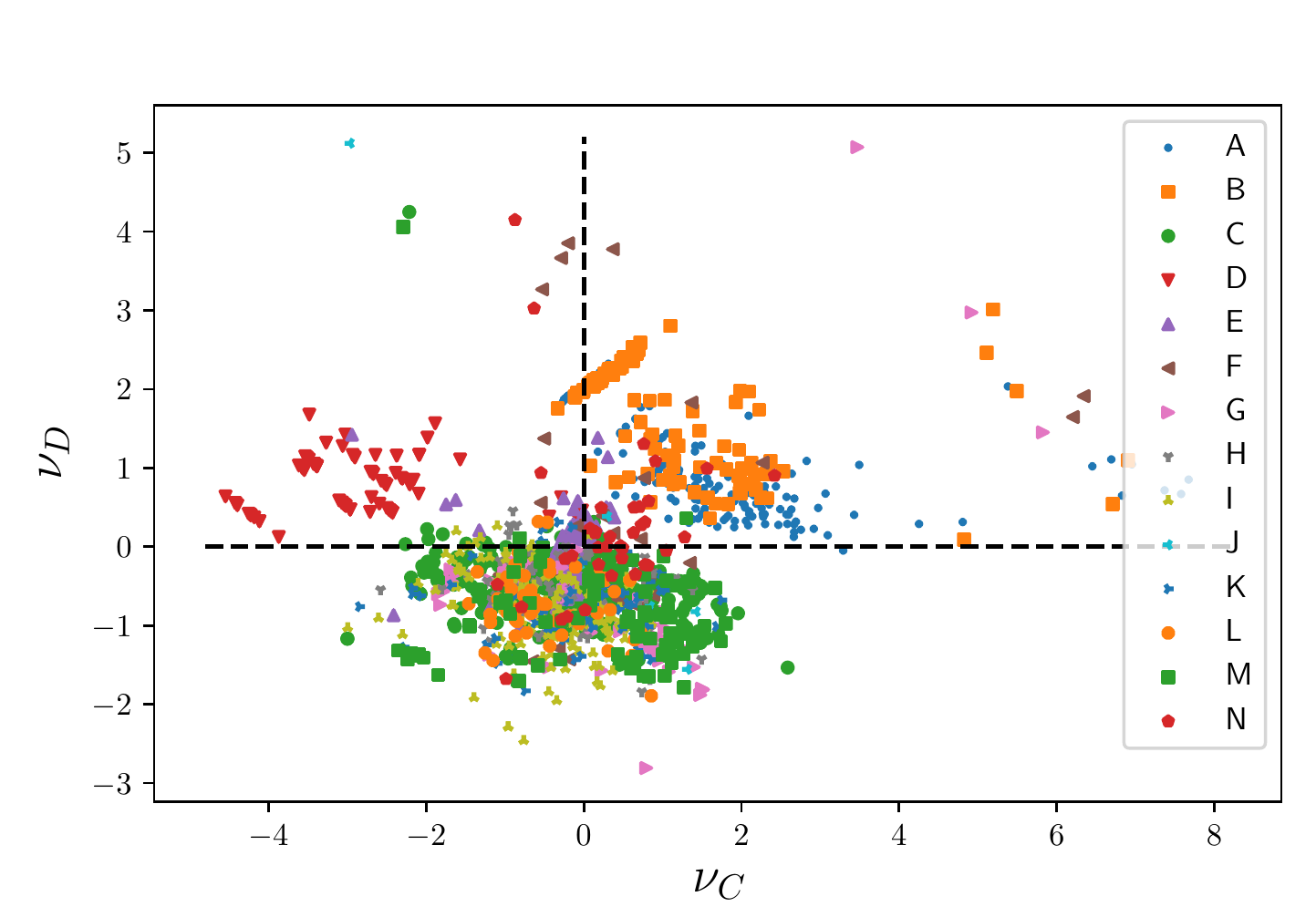}
    \caption{maximum \textit{a posteriori} estimates of the text propensities ($\NnuP_\Nj$, $\NnuQ_\Nj$), with their clustering class memberships. $\NnuP_\Nj$ drives the MAR effect and $\NnuQ_\Nj$ drives the \NMARacro one. 
    %Dashed lines are added for visual support.
    }
    \label{fig:asnt_nu_c_nu_d_grp}
\end{figure}

Going back to the model selected by ICL described in Section~\ref{sec:realdata}, we analyze the text propensities to be voted upon and to be positively perceived by nonvoters. These propensities  are encoded in the values of the latent variables $\NPb$ and $\NQb$. 
Figure~\ref{fig:asnt_nu_c_nu_d_grp} displays the scatter plot of $\NnuP_\Nj$ and $\NnuQ_\Nj$, the maximum \textit{a posteriori} estimates of $\NP_\Nj$ and $\NQ_\Nj$  under the variational distribution, for all voted texts. 
The abscissa $\NnuPb$ reflects the mobilization on the texts, with higher mobilization for higher values, and the ordinate  $\NnuQb$ represents the additional effect of mobilizing specifically  supporting voters.
The fourteen-cluster membership of texts (there is no obvious relevant classification for texts) is indicated by the plotting symbol.

Some relationship between missingness and membership to text classes emerge from this plot.
A first cluster of text appears in the positive quadrant, with texts mainly proposed by the government, categorized in text classes A and B.
A second cluster, smaller, on the upper left, is mainly formed by texts categorized in class D, voted positively by few voters. All the texts are related to the same law project regarding housing and were voted over a short period (06/03/2018 and 06/08/2018). 
The largest cluster,  on the lower part of the graph, gathers most of the remaining texts, that would have a tendency to be voted negatively by nonvoters. These texts were proposed by either the right-wing or left-wing opposition, and get little support from a vast majority of MPs.
%    \begin{itemize}
%        \item one, on the upper right on the graph, mainly formed by texts proposed by the government (texts from A and B) and are unsurprisingly voted positively as the MPs supporting the government outnumber the others
%        \item one, on the upper left on the graph, mainly formed by texts from class D are voted positively but are not notorious. All these texts are related to the same law project (housing) and are voted the two same days (06/08/2018 and 03/08/2018). 
%        \item the last one, on the lower part of the graph, gathers all the remaining texts that have a tendency to be voted negatively whatever their notoriety. These texts are proposed by either the right or left wing minorities and get little support from other MPs.
%    \end{itemize}
Note also that the small group of highly voted texts, on the right-hand side, is made of texts  belonging to six text classes. This reflects the fact that our model does not link the \NMARacro effect to the LBM memberships.

%
%%%%% MEME FIGURE QUE PRÉCÉDENTE MAIS AFFICHÉ SUR PLUSIEURS GRAPHS 
%
%\begin{figure}[H]
%    \includegraphics[width=1.\textwidth]{img/asnt_nu_c_nu_d_grp_splitted.pdf}
%    \caption{TODO}
%    \label{fig:asnt_nu_c_nu_d_splitted}
%\end{figure}

\end{document}